\documentclass[lettersize,journal]{IEEEtran}
\usepackage{amsmath,amsfonts}
\usepackage[utf8]{inputenc}
\usepackage{algorithmic}
\usepackage{algorithm}
\usepackage{array}
\usepackage[caption=false,font=normalsize,labelfont=sf,textfont=sf]{subfig}
\usepackage{textcomp}
\usepackage{stfloats}
\usepackage{url}
\usepackage{verbatim}
\usepackage{graphicx}
\usepackage{cite}
\usepackage{hyperref}
\usepackage{graphicx}
\usepackage{tikz}
\usepackage{xcolor}
\usepackage{hyperref}
\usepackage{xcolor}
\usepackage{scalerel}
\usepackage{tikz}
\usepackage{tikz}
\usepackage{multirow}

\usetikzlibrary{svg.path}
\definecolor{orcidlogocol}{HTML}{A6CE39}
\tikzset{
  orcidlogo/.pic={
    \fill[orcidlogocol] svg{M256,128c0,70.7-57.3,128-128,128C57.3,256,0,198.7,0,128C0,57.3,57.3,0,128,0C198.7,0,256,57.3,256,128z};
    \fill[white] svg{M86.3,186.2H70.9V79.1h15.4v48.4V186.2z}
                 svg{M108.9,79.1h41.6c39.6,0,57,28.3,57,53.6c0,27.5-21.5,53.6-56.8,53.6h-41.8V79.1z M124.3,172.4h24.5c34.9,0,42.9-26.5,42.9-39.7c0-21.5-13.7-39.7-43.7-39.7h-23.7V172.4z}
                 svg{M88.7,56.8c0,5.5-4.5,10.1-10.1,10.1c-5.6,0-10.1-4.6-10.1-10.1c0-5.6,4.5-10.1,10.1-10.1C84.2,46.7,88.7,51.3,88.7,56.8z};
  }
}

\newcommand\orcidicon[1]{\href{https://orcid.org/#1}{\mbox{\scalerel*{
\begin{tikzpicture}[yscale=-1,transform shape]
\pic{orcidlogo};
\end{tikzpicture}
}{|}}}}

\begin{document}

\title{Hierarchical Two-Stage Framework for Environment-Aware Long-Horizon Vessel Trajectory Prediction}

\author{Ganeshaaraj Gnanavel\orcidicon{0009-0001-9848-1097}\,, Sridha Sridharan\orcidicon{0000-0003-4316-9001}\,,~\IEEEmembership{Life Senior Member,~IEEE,} Clinton Fookes\orcidicon{0000-0002-8515-6324}\,,~\IEEEmembership{Senior Member,~IEEE}, and Tharindu Fernando\orcidicon{/0000-0002-6935-1816}\,,~\IEEEmembership{Member,~IEEE,}
        % <-this % stops a space
\thanks{Ganeshaaraj Gnanavel, Sridha Sridharan, Clinton Fookes, Tharindu Fernando are with the SAIVT Research Group, Queensland University of Technology, Brisbane, QLD 4000, Australia (email - g.gnanavel@hdr.qut.edu.au; t.warnakulasuriya@qut.edu.au; s.sridharan@qut.edu.au; c.fookes@qut.edu.au). (Corresponding author: G.Ganeshaaraj)}% <-this % stops a space
\thanks{}
}

\maketitle
\pagenumbering{gobble}

\begin{abstract}
Long-horizon vessel trajectory forecasting under real ocean conditions is critical for collision avoidance, traffic management, and route planning. However, achieving accurate predictions is challenging due to long-range temporal dependencies and dynamic environmental factors such as currents, wind, and waves. To address these issues, we propose a hierarchical two-stage framework that combines a coarse long-term predictor with a grid-aware short-term predictor through a hierarchical fusion mechanism. The short-term branch leverages a Spatio-Temporal Graph Transformer on discretized maritime cells to capture localized dynamics, while the long-term branch encodes overarching navigational intent. An integrated environmental module incorporates oceanographic parameters, including surface currents, wind vectors, and significant wave height, using cross-modal attention and feature-wise modulation for adaptive response to varying sea conditions. Additionally, a learnable Savitzky–Golay smoothing layer enhances temporal coherence in fused trajectories. We evaluate our approach on Australian Craft Tracking System (CTS) data from the North West region, aligned with Copernicus Marine Service products, using a 3-hour input and a 10-hour prediction horizon. Experimental results show that our framework outperforms the state-of-the-art by 25\% in Average Displacement Error (ADE) and 17\% in Final Displacement Error (FDE). Ablation studies further validate the contribution of each component.Source codes are available at \textcolor{blue}{\href{https://github.com/Ganeshaaraj-Ganesh/Hierarchical-Two-Stage-Framework-for-Environment-Aware-Long-Horizon-Vessel-Trajectory-Prediction/tree/main}{this GitHub repository}}
\end{abstract}


\section{Introduction}

Maritime transportation serves as a fundamental component of international trade and global logistics, accounting for over 80\% of global cargo movement. Its capacity to facilitate large-scale, long-distance, and cost-efficient transport has positioned it as a primary mode for international economic exchange. With the ongoing growth of the global shipping industry, the number of vessels operating in international waters continues to increase. This rise has contributed to more complex maritime traffic conditions, elevating the risk of collisions and accidents at sea \cite{szlapczynski2018determining}. These incidents pose risks to human life, cause delays in supply chains, and result in environmental harm. Vessel trajectory prediction has emerged as a key technical approach in addressing these challenges\cite{zhang2022vessel}. By estimating future vessel positions, trajectory prediction contributes to collision avoidance, supports port scheduling, informs route optimization, and assists search and rescue operations. As such, its integration into maritime safety systems is regarded as a necessary measure to improve operational efficiency and reduce risk across the maritime transport domain.

One major challenge in vessel trajectory prediction is long-term prediction. Long-term prediction is essential for strategic route planning, enabling proactive decision-making in maritime traffic management. It supports collision-avoidance systems by forecasting potential conflicts beyond the immediate horizon. Additionally, it enhances situational awareness for surveillance and security applications in open seas and coastal zones. Currently, most research in the vessel trajectory prediction domain focuses on short-term prediction\cite{jiang2024stmgf, xiao2022bidirectional}, resulting in only a limited number of studies addressing long-term forecasting. While existing deep learning models, such as Recurrent Neural Network (RNN) \cite{sherstinsky2020fundamentals} and Long Short Term Memory (LSTM)\cite{graves2012long}, demonstrate strong performance in short-term predictions, they encounter challenges in long-sequence time series forecasting (LSTF) due to difficulties in capturing long-term dependencies, leading to substantial prediction errors. To address these limitations, some recent studies have introduced various enhancements to the Transformer architecture aimed at improving accuracy in long-term forecasting\cite{xiong2024informer,qiang2023mstformer}. However, Transformer-based models often under-utilize short-term variations during long-term prediction due to their global attention focus. This limits their ability to capture local dynamics, contributing to cumulative forecast errors. While recent trajectory prediction methods incorporate attention mechanisms and contextual inputs, most approaches treat spatial scale and environmental forcing as independent factors. As a result, they lack a unified mechanism to capture the coupled and temporally evolving interaction between vessel dynamics and environmental conditions.

In this paper, we propose a hierarchically coupled two-stage framework for long-horizon vessel trajectory prediction that explicitly separates local motion refinement from global navigational intent and then recombines them through a learned fusion mechanism. This design enables the model to capture interactions between long-range navigational intent and short-term environmentally influenced deviations, which cannot be effectively modeled using single-stage or independently fused approaches. We use the term “grid-aware” to denote an explicit discretization of the maritime domain into spatial cells whose adjacency and occupancy features are encoded as a graph, enabling reasoning over cell-level interactions and transitions. Rather than directly applying STAR to raw vessel trajectories, the short-term module reformulates trajectory prediction as structured transitions over grid-defined maritime cells, enabling localized motion patterns to be modeled through spatial adjacency and transition dynamics. The long-term predictor captures broader navigational intent over extended horizons. A hierarchical fusion module adaptively arbitrates between the long-term and short-term branches, explicitly balancing global navigational intent with local trajectory corrections based on context-dependent confidence. To further improve the temporal smoothness of the final trajectory estimates, we implement a learnable Savitzky-Golay filter on top of the fused predictions.

Another key challenge in vessel trajectory prediction is modeling the influence of environmental factors. Environmental factors such as ocean current speeds, wind speeds, and wave height exert dynamic forces on vessels, altering their speed and direction beyond their intended course. These forces introduce nonlinearities in movement patterns that cannot be captured through GPS data alone. Incorporating such environmental data enables models to account for external influences that affect navigational behaviour. Although a few prior studies\cite{huang2022ea, xiao2024adaptive} have incorporated environmental factors into vessel trajectory prediction, they often consider a limited subset of these variables, neglecting the full range of relevant environmental influences. Moreover, these approaches typically fail to account for dynamically modeling the varying impact of spatial, temporal, and environmental factors on vessel movement, thereby limiting their predictive capability. In contrast to prior work that either uses limited environmental variables or treats them as auxiliary inputs, our framework integrates surface currents, wind vectors, and wave heights as temporally aligned conditioning signals, allowing environmental dynamics to directly influence trajectory evolution rather than being incorporated as static auxiliary inputs. This fusion mechanism enables the trajectory encoder to weight environmental and motion cues according to the current spatio-temporal context, rather than relying on static concatenation or fixed feature weighting. The feature modulation stage then conditions latent trajectory representations on the environmental context, enabling identical motion patterns to evolve differently under varying oceanographic conditions.

To the best of our knowledge, this is the first vessel trajectory forecasting framework to jointly combine hierarchical stage-wise prediction, grid-aware local refinement, and temporally aligned environmental conditioning within a unified long-horizon model. The proposed framework is therefore not defined by the introduction of new individual components, but by a structured integration that enables interaction-aware trajectory modeling across spatial scales and environmental conditions.

The main technical contributions of the proposed framework can be summarized as follows:

\begin{itemize}
\item We propose a hierarchical two-stage forecasting framework that explicitly separates long-horizon navigational intent from short-term local motion refinement, enabling the model to address both cumulative long-range error and localized trajectory variation.
\item We introduce a grid-aware short-term refinement branch and a learned hierarchical fusion mechanism that adaptively balances global trajectory consistency with local cell-level corrections.
\item We develop a unified environmental integration module that treats currents, wind, and wave height as temporally aligned conditioning signals, enabling environmental context to modulate trajectory representations rather than being appended as static auxiliary features.
\item We implement a learnable Savitzky-Golay filter to post-process the fused trajectory predictions, improving temporal consistency and smoothing without sacrificing structural accuracy.
\item Extensive experiments and ablation studies evaluate the contribution of the hierarchical structure, grid-aware refinement, environmental conditioning, and smoothing components under a 10-hour forecasting horizon.
\end{itemize}

\section{Related Work}

This section reviews prior research on data-driven vessel trajectory forecasting from three complementary perspectives. First, we summarize traditional, classical machine learning, and deep learning approaches to general vessel trajectory prediction. Next, we focus on methods explicitly designed for long-term vessel trajectory prediction. Finally, we discuss work that incorporates environmental factors into trajectory models and analyzes their impact on forecasting performance.

\subsection{Vessel Trajectory Prediction}
The traditional approaches to vessel trajectory prediction rely on empirical and mathematical models. Xu et al. \cite{xu2023long} developed a high-precision, long-period trajectory prediction system for oil tankers, utilizing density-based spatial clustering of applications with noise to analyze Automatic Identification System (AIS) data and identify a series of key points that denote essential navigation modes. Xiao et al. \cite{xiao2020big} introduced a novel model that integrates motion modeling and filtering processes. Srivastava et al. \cite{srivastava2023framework} introduced a lightweight, short-term forecasting model utilizing linear stationary models for predicting ship trajectories and detecting anomalies in real-time. However, these methodologies often suffer from high sensitivity to data quality and model complexity, which can amplify errors and lead to unreliable long-term predictions.

To mitigate these vulnerabilities, research has concentrated on classical machine-learning methodologies to precisely forecast vessel trajectories. Zhang et al. \cite{zhang2022short} developed a method to enhance the accuracy of ship position forecasts in maritime traffic engineering through the utilization of the k-nearest neighbors (KNN) algorithm. Sedagha et al. \cite{sedaghat2024deep} proposed a system framework for online marine traffic monitoring designed for the real-time surveillance of vessels on waterways and the forecasting of their future positions. Wei et al. \cite{wei2023three} devised a multi-objective heterogeneous integration method for predicting ship motion using a decomposition-reconstruction process and adaptive segmentation error correction. Xiao et al. \cite{lang2024physics} developed a machine learning model utilizing physical data to construct a gray box model (GBM) for predicting the speed of vessels traversing the ocean. Dong et al. \cite{dong2023math} proposed a novel mathematical data integration prediction (MDIP) model for forecasting the maneuvering movements of vessels. Nonetheless, classical machine learning techniques often struggle to adapt to the dynamic and heterogeneous behavior of vessels. As a result, generating accurate predictions remains challenging, limiting their effectiveness in real-world maritime environments.

Deep learning techniques proficiently leverage neural networks to extract high-dimensional features from extensive datasets and have been extensively utilized. Wu et al. \cite{wu2023ship} developed an integrated convolutional long short-term memory (LSTM) and sequence-to-sequence (Seq2Seq) model to enhance accuracy. Chen et al. \cite{chen2022automatic} introduced a novel framework for the precise acquisition and prediction of ship trajectories with a bidirectional LSTM (BiLSTM) model. Zhao et al. \cite{zhao2023ship} developed a model for forecasting ship trajectories by integrating a graph attention network with a long short-term memory (LSTM) network. Guo et al. \cite{guo2023toward} presented a multimodal data technique that integrates supplementary hidden states to independently delineate complex modes. Mehri et al. \cite{mehri2023context} introduced a contextually and data-driven approach for forecasting ship trajectories. Nonetheless, contemporary deep learning techniques face critical limitations; they often fail to capture complex contextual variations and effectively manage uncertainty, which undermines their reliability in dynamic, real-world environments.

\subsection{Long Term Vessel Trajectory Prediction}

Dong et al. \cite{dong2024attention} introduced a model utilizing an attention mechanism grounded in positional encoding to quantify ship movements, thereby extracting temporal correlations in trajectories and producing long-term predictions. Zhao et al. \cite{zhao2023large} presented an enhanced model that employs a retrospective window to capture temporal and spatial interdependence, therefore augmenting training efficiency and enhancing both short-term and long-term predictive accuracy. Transformer-based models have recently been utilized to capture long-range dependencies in AIS trajectories and enhance multi-step forecasting precision. Hunag et al. \cite{huang2022ea} introduced the Environment-Aware Vessel Trajectory Prediction Network (EA-VTP), which encodes navigational intent using CNN-derived vessel-density features integrated with recurrent sequence modeling, thereby enhancing long-horizon accuracy. Transformer-based models have recently been utilized to capture long-range dependencies in AIS trajectories and enhance multi-step forecasting precision. Nguyen et al. \cite{nguyen2024transformer} introduced a discrete, high-dimensional AIS representation and the TrAISformer, an enhanced Transformer that captures long-term correlations to forecast ship movements many hours in advance. Huang et al. \cite{huang2022tripleconvtransformer} developed TripleConvTransformer, integrating a streamlined Transformer with three convolutional pathways and amalgamating discretized meteorological fields with AIS to extract multi-scale motion signals and improve 90-minute forecasts. Building upon Informer\cite{zhou2021informer}, for extensive sequence time-series forecasting, Xiong et al. \cite{xiong2024informer} introduced Informer-TP, a model that consists of an encoder–decoder that utilises ProbSparse self-attention to effectively capture long-term dependencies for multi-ship, multi-step prediction. Qiang et al. \cite{qiang2023mstformer} presented MSTFormer, a motion-inspired spatial-temporal Transformer that enhances inputs with an Augmented Trajectory Matrix, employs multi-head dynamic-aware self-attention, and utilizes a knowledge-inspired loss to stabilize long-horizon forecasting, especially during turns. Notwithstanding these advancements, current attention- and Transformer-based methodologies continue to demonstrate numerous limitations in accurately anticipating long-term vessel trajectories. Numerous models are assessed mostly across short- to medium-range horizons (e.g., up to 90 minutes) or in limited traffic conditions, and their efficacy diminishes as the forecast horizon lengthens, especially during intricate maneuvers or in dense multi-ship interactions.

\subsection{Incorporation of Environmental factors}

Xiao et al. \cite{xiao2024adaptive} integrated AIS data with environmental parameters such as sea-wind speed/direction, visibility, and temperature. The authors integrated the environmental parameters utilizing parallel extractors and a fusion block to enhance short-term forecasts. Huang et al. \cite{huang2022tripleconvtransformer} present TripleConvTransformer, which concurrently learns AIS with discretized meteorological variables (gridded wind, waves/sea condition, precipitation/fog/low-light proxies, and temperature) to capture multi-scale effects along global routes. Bi et al. \cite{bi2024cnngru} developed a marine-fusion dataset and a model based on a convolutional neural network that incorporates AIS data along with wind direction and speed, wave height, and current direction and speed. In a complementary approach, in a context-aware pipeline, Mehri et al. \cite{mehri2023context} integrate environmental descriptors (wind speed/direction, wave height/direction, currents, and case-specific bathymetry) into AIS trajectories, conduct feature selection to reduce dimensionality, and employ a context-aware LSTM for forecasting. Guo et al. \cite{guo2025vessel} enhance Seq2Seq LSTMs by incorporating uncertainty estimation and an influence map, explicitly representing seawater depth (GEBCO bathymetry) and time-of-day/illumination conditions as a visibility proxy, with traffic context. Notwithstanding these advancements, significant challenges persist. These include the scarcity and heterogeneity of environmental observations, which limit the ability to capture fine-grained ocean conditions; mismatches in spatial and temporal scales between gridded environmental fields and vessel dynamics, which complicate integration; and the inherently nonstationary nature of maritime environments, where conditions evolve unpredictably over time.
\newcommand{\crosscircle}{%
\begin{tikzpicture}[baseline=-0.6ex] % baseline tweak so it sits nicely in text
    \fill[black] (0,0) circle (0.12cm);        % black filled circle
    \draw[line width=0.4mm, white] (-0.12cm,0) -- (0.12cm,0); % horizontal white bar
    \draw[line width=0.4mm, white] (0,-0.12cm) -- (0,0.12cm); % vertical white bar
\end{tikzpicture}%
}
\begin{figure*}[ht]
       \centering
       \includegraphics[width=\textwidth]
       {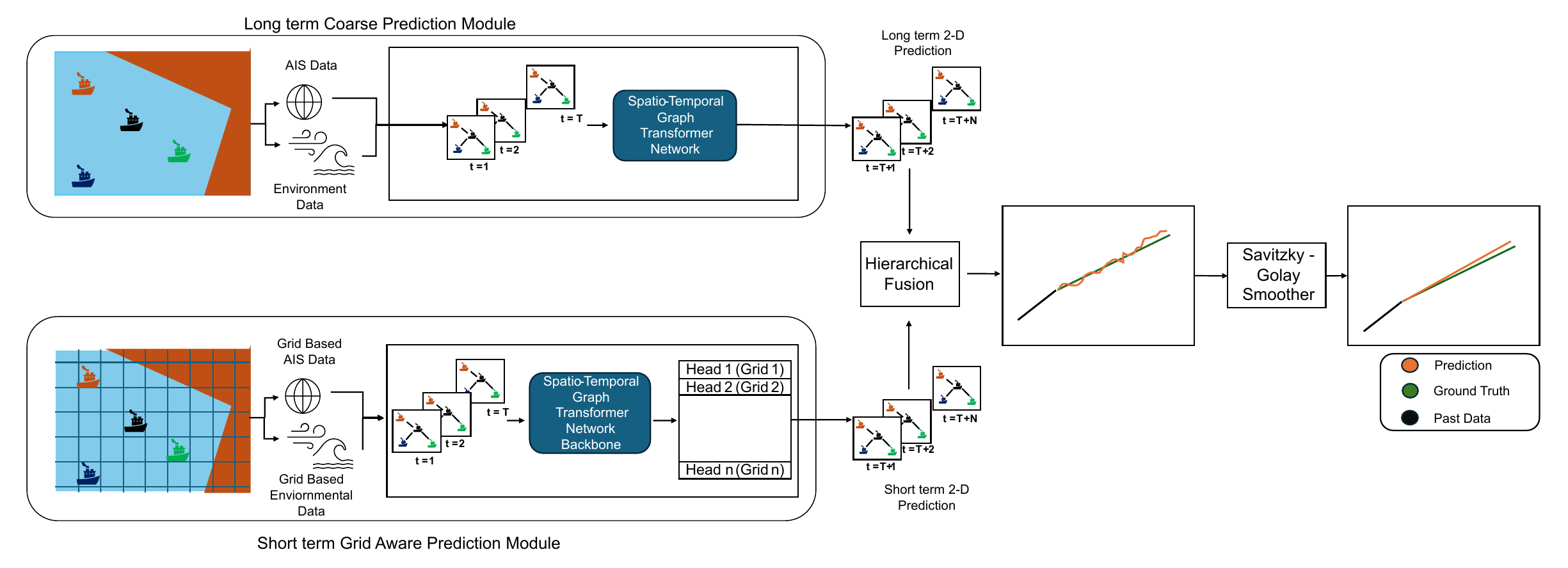}
       \captionsetup{justification=justified}
       \caption{{This figure presents a hierarchical two-stage framework designed for long-horizon, environment-aware vessel trajectory prediction. The architecture processes AIS kinematics and environmental data via parallel coarse long-term and grid-aware short-term branches. The signals are integrated through a hierarchical fusion module and refined using a learnable Savitzky-Golay smoothing layer to generate the final trajectory output.}}
       \label{fig:Pipeline}%
\end{figure*}

\section{Methodology}\label{meth}

This section presents our proposed framework for forecasting long-term vessel trajectories with environmental awareness. {Figure.~\ref{fig:Pipeline}} presents an overview. The methodology is structured around three primary components:

\begin{itemize}
    \item \textbf{Two-Stage Forecasting with Hierarchical Fusion}: A novel two-stage trajectory forecasting framework is proposed, consisting of distinct short-term and long-term prediction modules aimed at capturing fine-grained local variations and broader navigational intent, respectively. This study presents a novel grid-aware ensemble of predictors for short-term modeling, incorporating a hierarchical gating mechanism to integrate its output with the long-term predictor, thereby achieving localized accuracy alongside global trajectory coherence.

    \item \textbf{Unified Environmental Feature Module}: Oceanographic and meteorological variables are integrated via a unified feature module that combines trajectory dynamics with environmental context in a systematic and learnable format.

    \item \textbf{Learnable Smoothing Filter}: We implement a learnable filtering layer to improve temporal coherence while maintaining structural accuracy. This layer adaptively smooths the fused trajectories, minimizing high-frequency artifacts and preserving structural integrity.

\end{itemize}

Each of these components is described in detail in the subsections below.

\subsection{Problem Formulation}\label{meth_def}

Trajectory prediction constitutes a sequential prediction challenge. Given a series of past observed coordinates \(\{(x^n_t, y^n_t)\}_{n=1}^{N}\) for \(N\) vessels over the time steps \(t = -T_{\text{obs}} + 1, -T_{\text{obs}} + 2, \ldots, 0\), the objective is to predict the future locations \((\hat{x}_t, \hat{y}_t)\) of the vessel of interest for a constant period \(t = 1, 2, \ldots, T_{\text{pred}}\).

\subsection{Two-Stage Forecasting with Hierarchical Fusion}\label{meth_embedding_gen}

\subsubsection{Long Term Prediction}\label{meth_embedding_gen}

Long-Term Prediction: For the long-horizon branch, we adopt
\textsc{STAR}~\cite{yu2020spatio} as a spatio-temporal graph
encoder that maps observed trajectories, interaction graphs, and
environmental fields into a compact latent representation. We
emphasize that \textsc{STAR} is used to encode the past interaction
and environmental context, the resulting latent state is then passed
to our prediction head and downstream fusion module to produce
future displacements.

At each time step, we construct a directed interaction graph whose
vertices correspond to vessels and whose edges represent pairwise
interactions. For an edge from vessel \(i\) to vessel \(j\) at time \(t\),
we form an edge feature vector
\(\mathbf{e}_{ij,t}\) by concatenating spatial offsets
\((\Delta x_{ij,t}, \Delta y_{ij,t})\) with collocated environmental
variables: ocean currents \(\mathbf{c}_{ij,t}\), wind vectors
\(\mathbf{w}_{ij,t}\), and significant wave height \(H_{s,ij,t}\). Timestamps
\(t\) and positional encodings \(\mathbf{p}(t)\) are incorporated to index
the sequence.

The long-term encoder produces a time-indexed latent state
\(\mathbf{z}^{(L)}_{t} \in \mathbb{R}^{D}\), which summarizes the interaction
and environmental context up to time \(t\):
\begin{equation}
\mathbf{z}^{(L)}_{t}
= \Phi_{\star}\!\big(\mathbf{X}_{1:t},\, \mathbf{A}_{1:t},\, \mathbf{E}_{1:t}\big)
\in \mathbb{R}^{D},
\label{eq:long-encoder}
\end{equation}
where \(\Phi_{\star}\) denotes the \textsc{STAR} long-horizon spatio-temporal
encoder applied to trajectories \(\mathbf{X}\), interaction graphs
\(\mathbf{A}\), and edge/auxiliary features \(\mathbf{E}\) assembled from
\(\mathbf{e}_{ij,t}\), and \(\mathbf{z}^{(L)}_{t}\) is the resulting \(D\)-dimensional
long-term latent representation at time \(t\).

Within \textsc{STAR}, attention over vertex neighborhoods uses both
vertex states and edge features. Concretely, the edge features
\(\mathbf{e}_{ij,t}\) are concatenated with node states when forming the
keys and values in the attention mechanism, allowing the encoder
to weigh interactions based on spatial relationships and local
environmental context. This enables the model to prioritize
temporally distant but environmentally relevant interactions, which
is crucial when planning horizons extend beyond the local
kinematic window.

Both the long-term and short-term branches use a 48-dimensional
latent embedding. Accordingly, we set \(D = 48\) for the long-term
branch to match the latent size used in the short-term module (see
Eq.~(3)), enabling parameter sharing in the downstream fusion
module.

The decoder summarizes the future interaction context in the
encoded state \(\mathbf{z}^{(L)}_t\), from which we produce a 2-D
displacement prediction via a linear head:
\begin{equation}
\Delta\hat{\mathbf{s}}^{(L)}_{t}
= \mathbf{W}_{L}\,\mathbf{z}^{(L)}_{t} + \mathbf{b}_{L}
\in \mathbb{R}^{2},
\label{eq:long-head}
\end{equation}
where \(\Delta\hat{\mathbf{s}}^{(L)}_{t}
= (\Delta\hat{x}^{(L)}_{t}, \Delta\hat{y}^{(L)}_{t})\) denotes the
2-D displacement predicted by the long-term branch at time \(t\)
in the \((x,y)\) coordinate frame, and \(\mathbf{W}_{L} \in
\mathbb{R}^{2 \times 48}\) and \(\mathbf{b}_{L} \in \mathbb{R}^{2}\) are learned
parameters.

\subsubsection{short term grid aware prediction}\label{meth_embedding_SHORT}

\begin{figure*}[ht]
       \centering
       \includegraphics[width=\textwidth]{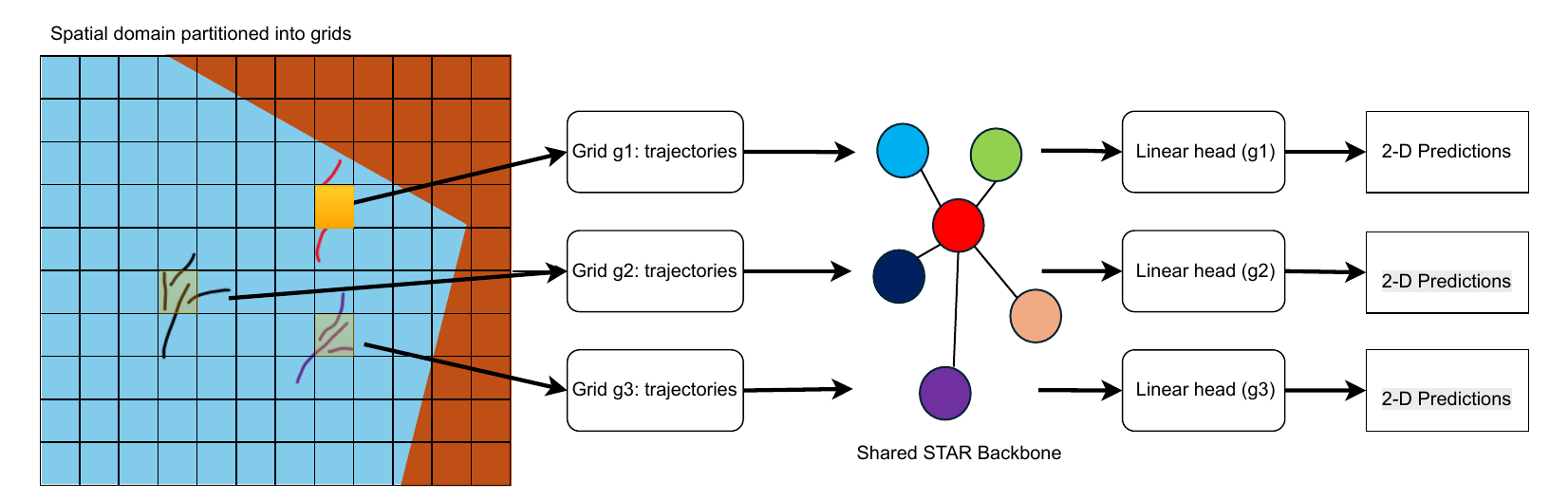}
    \caption{
        Illustration of the grid-aware short-term prediction module.
        The spatial domain is partitioned into 120 grids and vessel
        trajectories are grouped by their grid identifier. A single
        shared \textsc{STAR} backbone encodes the grid-wise trajectories
        into latent features, which are then passed through grid-specific
        linear heads to produce 2-D displacement predictions. The
        predictions are finally scattered back to the original vessel
        indices in the scene.
    }
    \label{fig:short-grid-star}
\end{figure*}

The short-term module employs STAR as the spatio-temporal framework and replicates the training configuration of the long-term branch. The spatial domain is divided into 120 grids, with each observation assigned a grid identifier. A single \textsc{STAR} backbone, shared across all spatial grids in the short-term branch, generates a latent trajectory representation, while a grid-specific linear head translates this representation into 2-D displacements, as illustrated
in Fig.~\ref{fig:short-grid-star}.

A single \textsc{STAR} backbone is associated with
120 grid-specific linear heads. In the forward pass, inputs are
grouped by grid, tensors are segmented along the vessel
dimension, and neighborhood tensors are sliced along both row
and column dimensions to preserve the graph structure. Each
unique grid in a batch is processed once by the backbone. The
corresponding linear head maps the resulting features to
2-D displacements, and the predictions are then scattered back
to their original indices.

Formally, letting \(g \in \{1,\dots,120\}\) denote the grid and \(t\)
a time step, the shared \textsc{STAR} encoder produces a 48-D latent
for each grid slice:
\begin{align}
\mathbf{z}_t^{(g)}
  &= \Phi_{\star}\!\left(
      \mathbf{X}_{1:t}^{(g)},\,
      \mathbf{A}_{1:t}^{(g)},\,
      \mathbf{E}_{1:t}^{(g)}
    \right)
  \in \mathbb{R}^{48}.
\end{align}
where \(\mathbf{X}_{1:t}^{(g)}\), \(\mathbf{A}_{1:t}^{(g)}\), and
\(\mathbf{E}_{1:t}^{(g)}\) denote the grid-restricted trajectories,
neighborhood graphs, and collocated environmental features,
respectively, following the notation of Eq.~\eqref{eq:long-encoder}
but restricted to grid \(g\).

This latent is further augmented by a grid-specific linear head into a 2-D displacement:
\begin{align}
\Delta \hat{\mathbf{s}}^{\,t}(g)
  &= W_g\, \mathbf{z}_t^{(g)} + \mathbf{b}_g
  \in \mathbb{R}^{2},
\end{align}
 $W_g \in \mathbb{R}^{2 \times 48}$ and $\mathbf{b}_g \in \mathbb{R}^{2}$ are learned per-grid parameters.

Distinct linear heads should be trained for each grid to effectively capture spatially varying motion fields and local interaction regimes, while also maintaining shared representation learning across the domain. The shared backbone facilitates the transfer of common trajectory and interaction structures, while head-specific parameters modify the final mapping to accommodate grid-level priors, traffic regulations, bathymetric constraints, and co-located environmental patterns. This separation minimizes interference among regions and facilitates the incremental adaptation of specific grids without necessitating retraining of the backbone. 

The inference process includes a single-backbone computation with a lightweight head selection based on the grid identifier, consistent with the two-stage fusion approach in which the short-term branch offers grid-aware adjustments to the long-term prediction. The details of the proposed fusion mechanism are illustrated below. 

\subsubsection{Hierarchical  Mechanism}\label{meth_fusion}

The final trajectory estimate in our framework is produced by a hierarchical fusion mechanism that adaptively combines the long-term and short-term predictors. This mechanism is implemented as a lightweight temporal CNN. After fusion inputs are constructed (Eq.~\ref{eq:fusion-input}), a temporal fusion operator $\Gamma(\cdot)$ maps the combined multi-branch sequence to the final fused displacement sequence. Operationally, $\Gamma(\cdot)$ consists of a dual-branch temporal encoder with both local and dilated temporal receptive fields, followed by a $1{\times}1$ convolutional gating layer. This $1{\times}1$ layer acts as a learned gate that balances long-horizon intent (from the long-term branch) against local, grid-aware dynamics (from the short-term branch).

To build the fusion input, we integrate the long-term (coarse) prediction with one or two short-term, grid-aware predictions. Each available short-term sequence is linearly upsampled to match the long-horizon length $T$ using $U_T(\cdot)$, where $U_T$ denotes a learnable linear upsampling operator. We denote by $\Delta \hat{\mathbf{s}}^{(L)}_t$ the displacement predicted by the long-term branch. We denote by $\Delta \hat{\mathbf{s}}^{(S_1,g)}_t$ the displacement predicted by the primary short-term, grid-aware module associated with grid $g$.

In some sequences, a vessel trajectory may traverse two distinct spatial grids during the input horizon or the forecast rollout (e.g., when crossing a grid boundary). In such cases, we instantiate a second short-term predictor head associated with the alternate grid, and obtain $\Delta \hat{\mathbf{s}}^{(S_2,g')}_t$, where $g'$ denotes the second grid. This provides an additional locally conditioned short-term signal that reflects the vessel's motion dynamics and priors in that neighboring grid region. If the trajectory remains within a single grid, the $(S_2,g')$ term is not present, and only the primary short-term branch $(S_1,g)$ contributes to fusion.

% Fusion input construction (omit the third term if there is no second grid-conditioned branch)
\begin{equation}
\begin{aligned}
\mathbf{f}_t
= \Big[
&\underbrace{\Delta \hat{\mathbf{s}}^{(L)}_{t}}_{\text{long-term (coarse)}}\;;\;
\underbrace{\big(U_T\, \Delta \hat{\mathbf{s}}^{(S_1,g)}\big)_t}_{\text{short-term (grid $g$)}}\;;\\
&\underbrace{\big(U_T\, \Delta \hat{\mathbf{s}}^{(S_2,g')}\big)_t}_{\text{short-term (grid $g'$, if applicable)}}
\Big],
\quad t = 1{:}T.
\label{eq:fusion-input}
\end{aligned}
\end{equation}

The fused displacement sequence for the current grid context is then obtained as:
\begin{equation}
\Delta \hat{\mathbf{s}}^{(F,g)}_{1:T}
= \Gamma\!\big(\mathbf{f}_{1:T}\big),
\label{eq:fusion-output}
\end{equation}

where $\Delta \hat{\mathbf{s}}^{(F,g)}_{1:T}$ is the fused 2-D displacement sequence. When only a single grid is active (i.e., the vessel does not cross a grid boundary), $\mathbf{f}_t$ excludes the $(S_2,g')$ branch and fusion reduces to combining the long-term branch with a single short-term source. Intuitively, the two temporal branches within $\Gamma(\cdot)$ specialize in different time scales: one branch (dilation $d{=}1$) captures short-horizon corrections, while the other (dilation $d{=}4$) captures longer-range temporal structure. The subsequent $1{\times}1$ convolution then serves as a hierarchical gate that adaptively blends these signals into a single fused trajectory.

Finally, we define the temporal fusion operator $\Gamma(\cdot)$:
\begin{equation}
\Gamma(\cdot)
= \mathrm{Conv}_{1\times 1}\!\left(
  \sigma\!\big(\mathrm{Conv}_{k=3,\,d=1}(\cdot)\big)
  \;\Vert\;
  \sigma\!\big(\mathrm{Conv}_{k=3,\,d=4}(\cdot)\big)
\right),
\label{eq:gamma-def}
\end{equation}

where $\mathrm{Conv}_{k=3,\,d=1}$ is a 1D temporal convolution with kernel size 3 and dilation 1 (local temporal context), $\mathrm{Conv}_{k=3,\,d=4}$ is a parallel 1D temporal convolution with kernel size 3 and dilation 4 (broader temporal context), $\sigma$ is the GELU activation, and ``$\Vert$'' denotes channel-wise concatenation of the two branches. The subsequent $\mathrm{Conv}_{1\times 1}$ is a pointwise temporal convolution that blends the concatenated features into the fused displacement.

This hierarchical fusion mechanism therefore adapts to spatial transitions: when a vessel remains within a single grid, the fusion relies on one grid-aware short-term predictor; when the vessel crosses into a second grid, the fusion incorporates both grid-specific short-term predictors, allowing the model to reconcile distinct local motion priors while preserving long-horizon structure from the coarse predictor.

\subsection{Unified Environmental Integration Module}\label{meth_embedding_gen}

In the next stage of the fusion, we integrate the environmental parameters. 
The Unified Environmental Integration Module operates on three feature streams at each timestep $t$: a temporal motion encoding $\mathbf{h}_t \in \mathbb{R}^{32}$, a spatial/context encoding $\mathbf{u}_t \in \mathbb{R}^{32}$, and an environmental encoding $\mathbf{e}_t \in \mathbb{R}^{32}$. These features are produced during node/edge feature construction in both the long-term and short-term prediction branches. Specifically,
\begin{itemize}
    \item $\mathbf{h}_t$ is derived from the trajectory encoder associated with the corresponding prediction branch (long-term or short-term) and captures the vessel's recent kinematic state.
    \item $\mathbf{u}_t$ encodes spatial and interaction context local to the vessel. This includes neighborhood structure from the interaction graph $\mathbf{A}_{1:t}$, grid-cell occupancy, and vessel-to-vessel relational cues.
    \item $\mathbf{e}_t$ encodes the collocated environmental/oceanographic conditions at time $t$, such as sea-surface current vectors $\mathbf{c}$, wind vectors $\mathbf{w}$, and significant wave height $H_s$.
\end{itemize}

We then fuse these modalities in two steps.

Step 1: Cross-attention alignment.
A cross-attention block uses $\mathbf{h}_t$ as the query and $[\mathbf{u}_t; \mathbf{e}_t]$ as the keys/values to produce an aligned representation:
\begin{equation}
    \tilde{\mathbf{h}}_t
    =
    \mathrm{MLP}\!\Big(
    \mathrm{Attn}\big(
        \mathbf{h}_t,\,
        [\mathbf{u}_t; \mathbf{e}_t],\,
        [\mathbf{u}_t; \mathbf{e}_t]
    \big)
    \Big)
    \in \mathbb{R}^{32}.
    \label{eq:env-cross-attn}
\end{equation}
This allows the temporal motion pathway to selectively attend to spatial and environmental factors that are most relevant at that timestep.

Step 2: FiLM-based modulation.
We next apply feature-wise linear modulation (FiLM) using a lower-level environmental signal $\mathbf{r}_t \in \mathbb{R}^{E}$. The FiLM layer generates per-channel scale and shift terms $\gamma(\mathbf{r}_t)$ and $\beta(\mathbf{r}_t)$, which then modulate $\tilde{\mathbf{h}}_t$:
\begin{equation}
    \bar{\mathbf{h}}_t
    =
    \big(1 + \gamma(\mathbf{r}_t)\big)
    \odot
    \tilde{\mathbf{h}}_t
    +
    \beta(\mathbf{r}_t),
    \label{eq:film-mod}
\end{equation}
where,
\begin{equation}
    \left[
    \gamma(\cdot), \beta(\cdot)
    \right]
    =
    \mathbf{W}_{\mathrm{film}}\, \mathbf{r}_t
    +
    \mathbf{b}_{\mathrm{film}}
    \in \mathbb{R}^{64}.
    \label{eq:film-params}
\end{equation}
Here, $\gamma(\mathbf{r}_t)$ and $\beta(\mathbf{r}_t)$ provide learnable, per-channel scaling and shifting of $\tilde{\mathbf{h}}_t$. This step adaptively conditions the motion representation on instantaneous environmental forcing, rather than simply concatenating environment features.

The resulting $\bar{\mathbf{h}}_t$ is then forwarded to the refinement encoder and prediction head of the corresponding branch (long-term or short-term). In other words, this unified integration block is applied consistently wherever node/edge features are formed for prediction, and it provides a principled mechanism for injecting oceanographic context into both stages of the forecasting pipeline.

This design enables dynamic adaptation to diverse environmental and spatial conditions that fixed feature fusion cannot accommodate. Traditional concatenation presumes linear and static interactions between modalities, potentially restricting generalization across varied physical or oceanographic contexts. The proposed module incorporates hierarchical conditioning, utilizing attention for context selection and employing FiLM to enhance feature influence via learnable modulation parameters. This structure enables the model to interpret environmental cues, including current flow, wind direction, and spatial constraints, while adjusting its internal state consistently over time.

\subsection{Learnable smoothing filter}\label{meth_embedding_gen}

The Savitzky-Golay (SG) filter \cite{schafer2011savitzky} was chosen for the post-processing of predicted trajectories due to its ability to maintain structural features while minimizing temporal noise. The SG filter, in contrast to simple averaging, conducts a local polynomial regression within a moving window, yielding smoothed outputs that preserve curvature information. This property is significant in contexts where predictions pertain to motion continuity and changes in direction. The model utilizes various SG filters with distinct window sizes to perform smoothing operations across multiple temporal scales. Each filter is implemented as a fixed one-dimensional convolution, utilizing coefficients derived from the corresponding polynomial basis, thereby ensuring compatibility with gradient-based optimization and backpropagation. Concretely, for window \(w\) with fixed Savitzky--Golay coefficients \(c_{\tau}(w)\), the depth-wise convolution acting on the 2-D displacements \(\Delta \hat{s}(F,g)\) is defined as,
\begin{equation}
\begin{aligned}
(\mathrm{SG}_{w}\!\star\!\Delta \hat{s})(F,g)_t
&= \sum_{\tau=-(w-1)/2}^{(w-1)/2} c_{\tau}(w)\,
      \Delta \hat{s}_{t-\tau}(F,g),\\[-2pt]
&\text{for } t=1,\dots,T .
\end{aligned}
\end{equation}

The learnable smoother aggregates the outputs of these filters through a softmax-weighted combination, allowing the network to determine which temporal window contributes most to the final output. In our formulation, the smoothed trajectory \(\Delta \hat{s}(S,g)\) is obtained by
\begin{equation}
\begin{aligned}
\Delta \hat{s}_{1:T}(S,g)
&= \sum_{k=1}^{K}\pi_k\,
      \bigl(\mathrm{SG}_{w_k}\!\star\!\Delta \hat{s}(F,g)\bigr)_{1:T},\\
\pi&=\operatorname{softmax}(\theta),\qquad
\theta\in\mathbb{R}^{K},
\end{aligned}
\end{equation}
with mixture weights \(\theta\) shared across channels and time. The learnable weights dynamically adjust the balance between short- and long-window smoothing throughout training, enhancing temporal consistency while preserving geometric structure. This method allows the model to incorporate temporal filtering directly into the learning process instead of implementing it as a separate post-processing step. This design aims to enhance dynamic stability in trajectory evolution, minimize noise accumulation from multi-stage predictions, and ensure that the final trajectories conform to realistic motion dynamics.

\section{Experiments}

In this section, we describe the experimental setup and present our findings. We evaluate the proposed framework across multiple datasets, demonstrate the effectiveness of incorporating environmental features, and analyze its performance using well-established evaluation metrics.

\subsection{Datasets}

\begin{figure}[h!]
    \centering
    \includegraphics[width=0.75\linewidth]{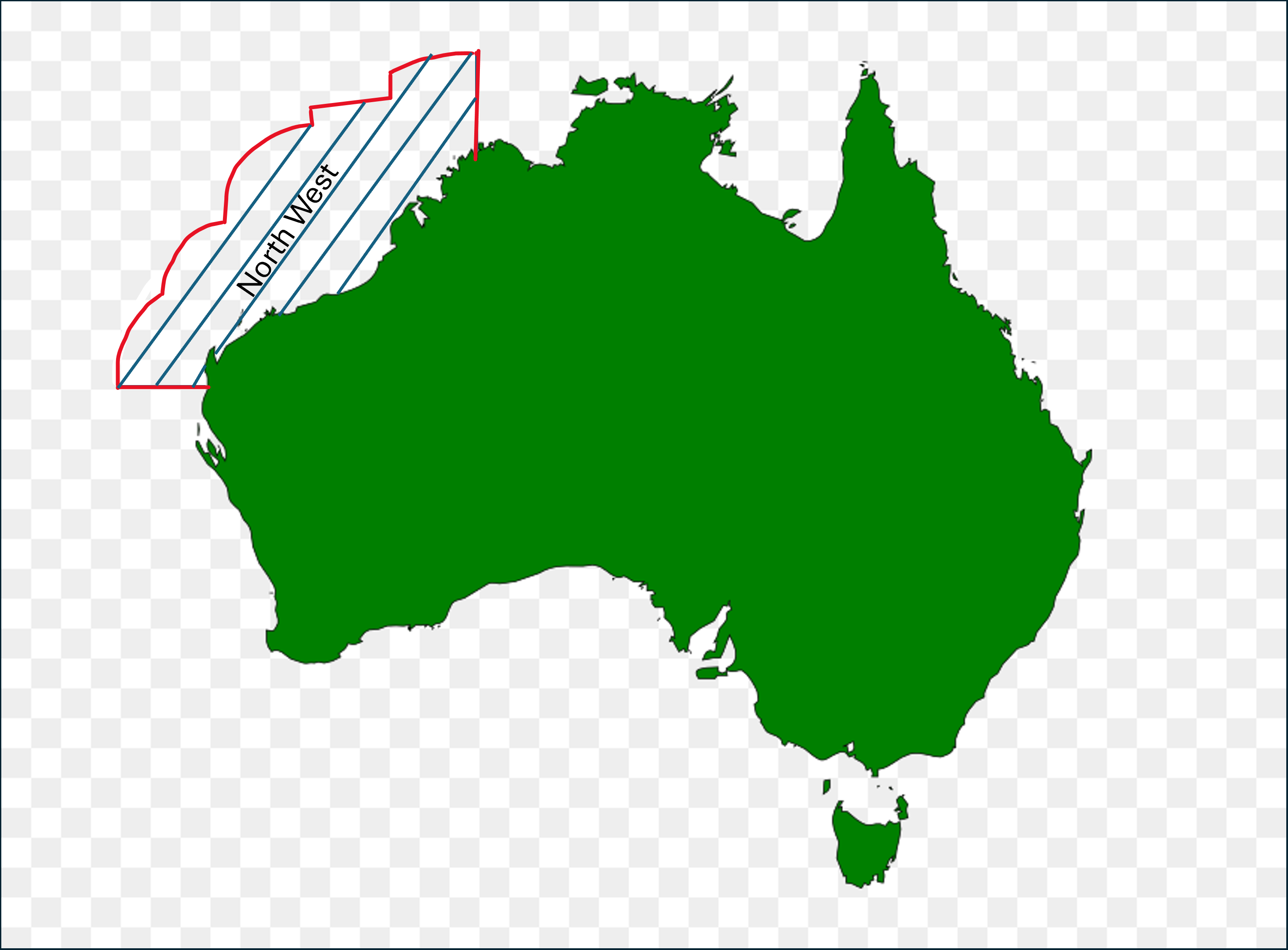}
    \caption{Map showing the areas of interest}
    \label{fig:AIS}
\end{figure}

In our experiments, we have utilised the Craft Tracking System (CTS) data\footnote{\url{https://www.amsa.gov.au/vessels-operators/maritime-safety-tracking/craft-tracking-system}}. This data was provided by the Australian Maritime Safety Authority. CTS gathers vessel traffic data from a variety of sources, including both terrestrial and satellite shipborne Automatic Identification System (AIS) data. The AIS is a transponder-based tracking and navigation system that shares static and dynamic vessel information. The transponder gathers and transmits vessel data via VHF radio transmissions. In our experiments, we focused on the North West region of Australia. The evaluation is performed in a single geographic location, a decision aligned with previous studies\cite{qiang2023mstformer,zhou2021informer,liu2019vtp}, as the chosen area encompasses a broad and representative spatial scope. The North West Area is bounded by longitudes $109.981^\circ\mathrm{E}$ to $125.335^\circ\mathrm{E}$ and latitudes $22.985^\circ\mathrm{S}$ to $11.515^\circ\mathrm{S}$. In {Fig.~\ref{fig:AIS}}, this area of interest is highlighted. Throughout the development of the framework and quantitative evaluation, AIS data from 01/01/2024 to 31/08/2024 served as training data, 01/09/2024 to 30/09/2024 as validation data, and 01/10/2024 to 31/10/2024 as testing data.

% \begin{table}[H]
% \centering
% \resizebox{\columnwidth}{!}{%
%     \begin{tabular}{|l|p{10cm}|c|}
%         \hline
%         \textbf{Features} & \textbf{Description} & \textbf{Unit} \\
%         \hline
%         MMSI & Maritime Mobile Service Identity number. \newline
%         - 9 digit number, unique for all vessels. \newline
%         - First 3 digits represent the vessel’s flag of registration. & unitless \\
%         \hline
%         Unixtime & Timestamp of AIS message \newline
%         - Unix format represents number of seconds that have elapsed since January 1, 1970. & seconds \\
%         \hline
%         Latitude & Geographic coordinate (WGS84 format) \newline
%         - Measured in decimal degrees [$-90^\circ$, $90^\circ$] & degrees \\
%         \hline
%         Longitude & Geographic coordinate (WGS84 format) \newline
%         - Measured in decimal degrees [$-180^\circ$, $180^\circ$] & degrees \\
%         \hline
%         Speed over ground (SOG) & Vessel’s speed relative to the earth’s surface. \newline
%         - Measured in knots (nautical miles/hour) & knots \\
%         \hline
%         Course over ground (COG) & Vessel’s course over the earth’s surface, measured clockwise relative to true north. \newline
%         - Expressed in degrees ($0^\circ$ to $360^\circ$) & degrees \\
%         \hline
%     \end{tabular}
%     }
%     \caption{Identified relevant features of the AIS dataset}
%     \label{tab:ais_features}
% \end{table}

We utilized weather and environmental data from the \emph{Copernicus Marine Service}\footnote{\url{https://doi.org/10.48670/moi-00016}} to obtain current environmental conditions over the identical spatial and temporal parameters. We utilized (i) sea-surface current velocity, (ii) sea-surface wind velocity, and (iii) significant wave height. Environmental factors were temporally aligned and spatially matched with AIS observations to provide a unified dataset for subsequent modeling and assessment.

To assess long-horizon forecasting performance, we formulate a multi-step prediction task with an input window of \(L=3\,\mathrm{h}\) and a forecasting horizon of \(H=10\,\mathrm{h}\). Models ingest the preceding \(L\) hours of observations and predict the subsequent \(H\) hours. This protocol is applied uniformly across all experiments and baseline models to ensure fair, like-for-like comparison of our method's performance.

To mitigate concerns about region-specific inductive biases and assess the framework's adaptability to various marine traffic patterns, we performed an extra cross-regional spatial study. We dynamically partitioned the domain according to historical AIS ping density, training the shared backbone exclusively on low-traffic grids and evaluating its generalization on high-density routes. Extensive information regarding the approach of this density-based spatial split, along with its associated quantitative outcomes, is available in the \ref{Cross-Regional Spatial Generalization Analysis} .

\subsection{Environmental Data Interpolation Method}\label{Environmental Data Interpolation Method}

To address the spatiotemporal resolution discrepancies between the high-frequency Automatic Identification System (AIS) and the coarse environmental grids from the Copernicus Marine Service, a rigorous spatiotemporal interpolation method is employed. For each AIS observation at a specific coordinate $(x_t, y_t)$ and time $t$, the framework employs the surrounding environmental grid nodes to compute a mathematically weighted, continuous feature space. Bilinear spatial interpolation utilizes the four nearest geographical grid corners to ascertain the localized environmental condition. This spatial mapping is followed by linear temporal interpolation between consecutive weather prediction time-steps to align accurately with the timestamp of the vessel's AIS ping.This technique is utilized to preserve the physical authenticity of the maritime environment. This technique establishes a mathematical relationship between coarse grid points in both spatial and temporal dimensions, enabling the synthesis of smooth, high-fidelity environmental gradients, including progressively increasing wind vectors and varying significant wave heights. This ensures that substantial meteorological changes are consistently distributed across the standard 15-minute AIS intervals, accurately reflecting the ongoing, continuous nature of the dynamic oceanographic forces experienced by a vessel at sea.

\subsection{Evaluation Metrics}

Consistent with \cite{jiang2024stmgf}, we employ average displacement error (ADE) and final displacement error (FDE) for performance evaluation. ADE computes the average Euclidean distance between the predicted trajectory and the ground truth over the entire prediction horizon. FDE measures the Euclidean distance at the final forecast time step between the predicted endpoint and the ground-truth endpoint.

\subsection{Quantitative Evaluation}

This section assesses our two-stage trajectory forecasting framework by comparing its long-term forecasting outcomes with several strong, widely adopted sequence-modeling baselines. We select LSTM~\cite{alahi2016social}, ConvLSTM~\cite{alahi2016social}, and the Transformer~\cite{giuliari2021transformer} as strong, widely adopted sequence-modeling baselines that represent recurrent, convolutional, and attention-based architectures for multistep time-series forecasting. In addition, we include MSTFormer~\cite{qiang2023mstformer}, Informer~\cite{xiong2024informer}, and a third-order Markov modell~\cite{liu2019vtp} as representative state-of-the-art and classical approaches for long-horizon vessel trajectory prediction, enabling a fair and comprehensive comparison across both traditional and modern paradigms. Our model has superior ADE and FDE compared to all baselines, as illustrated in {Table~\ref{table:comparison_vessel}}. ADE is enhanced by 25\% and FDE by 17\% compared to Informer's prior state-of-the-art performance. These advancements signify an enhancement in long-term trajectory forecasting and validate the two-stage architecture. We present findings for the two versions of the proposed framework. The two-stage trajectory forecasting framework, devoid of environmental inputs, outperforms previous state-of-the-art baselines, and the incorporation of environmental data results in further enhancements.

To specifically examine the safety and reliability limits essential for marine collision avoidance, we additionally assess the error distribution within the test set by presenting the 75th, 90th, and 95th percentiles. Although mean error metrics are vulnerable to infrequent outliers, the 95th percentile effectively constrains the worst-case tail risk of our deterministic forecasts. Our proposed framework restricts the maximum tail-risk error, capping the 95th percentile FDE at 5.46 kilometers. Conversely, the previous state-of-the-art Informer has a substantial decline in performance in edge instances, demonstrating a worst-case FDE of 8.57 km. 

% \begin{table*}[htbp]
% \caption{ADE / FDE (km) on the test set including error distribution percentiles. Lower values are better. Bold fonts represent the best results.}
% \centering
% \resizebox{\textwidth}{!}{%
% \begin{tabular}{|l|c|c|c|c|c|c|c|c|c|c|}
% \hline
% \multirow{\textbf{Method}} & \multicolumn{5}{c|}{\textbf{ADE (km)}} & \multicolumn{5}{c|}{\textbf{FDE (km)}} \\ \cline{2-11} 
%  & \textbf{Mean} & \textbf{Median} & \textbf{75th Pctl} & \textbf{90th Pctl} & \textbf{95th Pctl} & \textbf{Mean} & \textbf{Median} & \textbf{75th Pctl} & \textbf{90th Pctl} & \textbf{95th Pctl} \\ \hline
% LSTM & 7.34 & 6.54 & 8.83 & 11.56 & 13.28 & 12.60 & 11.27 & 15.18 & 19.84 & 23.59 \\ \hline
% Transformer & 4.28 & 3.86 & 5.13 & 6.87 & 8.04 & 5.89 & 5.17 & 7.16 & 9.47 & 11.29 \\ \hline
% ConvLSTM & 5.56 & 4.96 & 6.73 & 8.85 & 10.27 & 8.60 & 7.68 & 10.36 & 13.59 & 15.86 \\ \hline
% MSTFormer & 3.48 & 3.08 & 4.19 & 5.46 & 6.37 & 4.21 & 3.67 & 4.86 & 6.48 & 8.18 \\ \hline
% 3-Order Markov & 3.31 & 2.98 & 3.99 & 5.16 & 5.97 & 4.71 & 4.17 & 5.67 & 7.48 & 8.96 \\ \hline
% Informer & 2.98 & 2.69 & 3.66 & 4.68 & 5.46 & 4.43 & 3.97 & 5.26 & 6.87 & 8.57 \\ \hline
% Ours (W/out Env features) & 2.65 & 2.39 & 3.19 & 4.17 & 4.86 & 4.02 & 3.49 & 4.57 & 5.87 & 7.26 \\ \hline
% \textbf{Ours (With Env Features)} & \textbf{2.23} & \textbf{2.04} & \textbf{2.69} & \textbf{3.36} & \textbf{3.87} & \textbf{3.68} & \textbf{3.09} & \textbf{4.07} & \textbf{4.97} & \textbf{5.56} \\ \hline
% \end{tabular}%
% }
% \label{table:comparison_vessel}
% \end{table*}

\begin{table*}[htbp]
\caption{ADE / FDE (km) on the test set including error distribution percentiles. Lower values are better. Bold fonts represent the best results.}
\centering
\resizebox{\textwidth}{!}{%
\begin{tabular}{|l|c|c|c|c|c|c|c|c|}
\hline
\multirow{2}{*}{\textbf{Method}} & \multicolumn{4}{c|}{\textbf{ADE (km)}} & \multicolumn{4}{c|}{\textbf{FDE (km)}} \\ \cline{2-9} 
 & \textbf{Mean} & \textbf{75th Pctl} & \textbf{90th Pctl} & \textbf{95th Pctl} & \textbf{Mean} & \textbf{75th Pctl} & \textbf{90th Pctl} & \textbf{95th Pctl} \\ \hline
LSTM & 7.34 & 8.83 & 11.56 & 13.28 & 12.60 & 15.18 & 19.84 & 23.59 \\ \hline
Transformer & 4.28 & 5.13 & 6.87 & 8.04 & 5.89 & 7.16 & 9.47 & 11.29 \\ \hline
ConvLSTM & 5.56 & 6.73 & 8.85 & 10.27 & 8.60 & 10.36 & 13.59 & 15.86 \\ \hline
MSTFormer & 3.48 & 4.19 & 5.46 & 6.37 & 4.21 & 4.86 & 6.48 & 8.18 \\ \hline
3-Order Markov & 3.31 & 3.99 & 5.16 & 5.97 & 4.71 & 5.67 & 7.48 & 8.96 \\ \hline
Informer & 2.98 & 3.66 & 4.68 & 5.46 & 4.43 & 5.26 & 6.87 & 8.57 \\ \hline
Ours (W/out Env features) & 2.65 & 3.19 & 4.17 & 4.86 & 4.02 & 4.57 & 5.87 & 7.26 \\ \hline
\textbf{Ours (With Env Features)} & \textbf{2.23} & \textbf{2.69} & \textbf{3.36} & \textbf{3.87} & \textbf{3.68} & \textbf{4.07} & \textbf{4.97} & \textbf{5.56} \\ \hline
\end{tabular}
}
\label{table:comparison_vessel}
\end{table*}

\section{Ablation Study and Discussions}

\subsection{Analysis on the effect of spatiotemporal alignment frequency}

This phase of the ablation study investigates the impact of spatiotemporal alignment frequency on prediction accuracy by assessing the performance of the proposed framework against the coarse temporal resolutions of the available environmental data. The two assessed versions are: (i) \textit{Env-Coarse}, where environmental variables are sampled at their original, coarse resolution; and (ii) \textit{Ours (15-min)}, where all environmental variables are interpolated to a consistent 15-minute resolution to align with the vessel's AIS frequency. The 15-minute bin is chosen as it accurately corresponds to the vessel's standard 15-minute AIS intervals and environmental gradients.

This analysis is crucial for evaluating the influence of environmental context accuracy on the 10-hour prediction horizon, particularly with the noted disparities in spatial and temporal scales between gridded meteorological fields and high-frequency vessel dynamics. Table II demonstrates that improving the temporal resolution of environmental signals through our 15-minute spatiotemporal interpolation reliably reduces both ADE and FDE. The \textit{Env-Coarse} baseline provides overall background; however, its coarse resolution fails to capture the dynamic fluctuations in sea state between sporadic measurements, leading to heightened cumulative displacement errors.

The proposed 15-minute alignment enhances precision by providing high-fidelity environmental gradients that correspond exactly with the vessel's 15-minute kinematic states. This alignment stabilizes cross-modal attention and feature-wise linear modulation (FiLM) mechanisms by eliminating non-physical discontinuities in ambient vectors. The Spatio-Temporal Graph Transformer backbone improves trajectory prediction precision by capturing immediate environmental influences via the temporal motion channel, so more effectively linking localized sea conditions with long-term navigational objectives.

\begin{table}[h]
\centering
\caption{Analysis on the importance of spatiotemporal alignment frequency.}
\begin{tabular}{|l|c|c|}
\hline
Resolution & Env-Coarse (Raw) & Ours (15-min) \\ \hline
ADE (km) & 2.51 & \textbf{2.23} \\ \hline
FDE (km) & 3.95 & \textbf{3.68} \\ \hline
\end{tabular}
\label{tab:alignment}
\end{table}

\subsection{Analysis on the effect of learnable Savitzky-Golay filter}\label{ablation_cluster}

\begin{figure*}[t]
    \centering

    % Sub-figures stacked vertically
    \subfloat[]{\includegraphics[width=8.5cm]{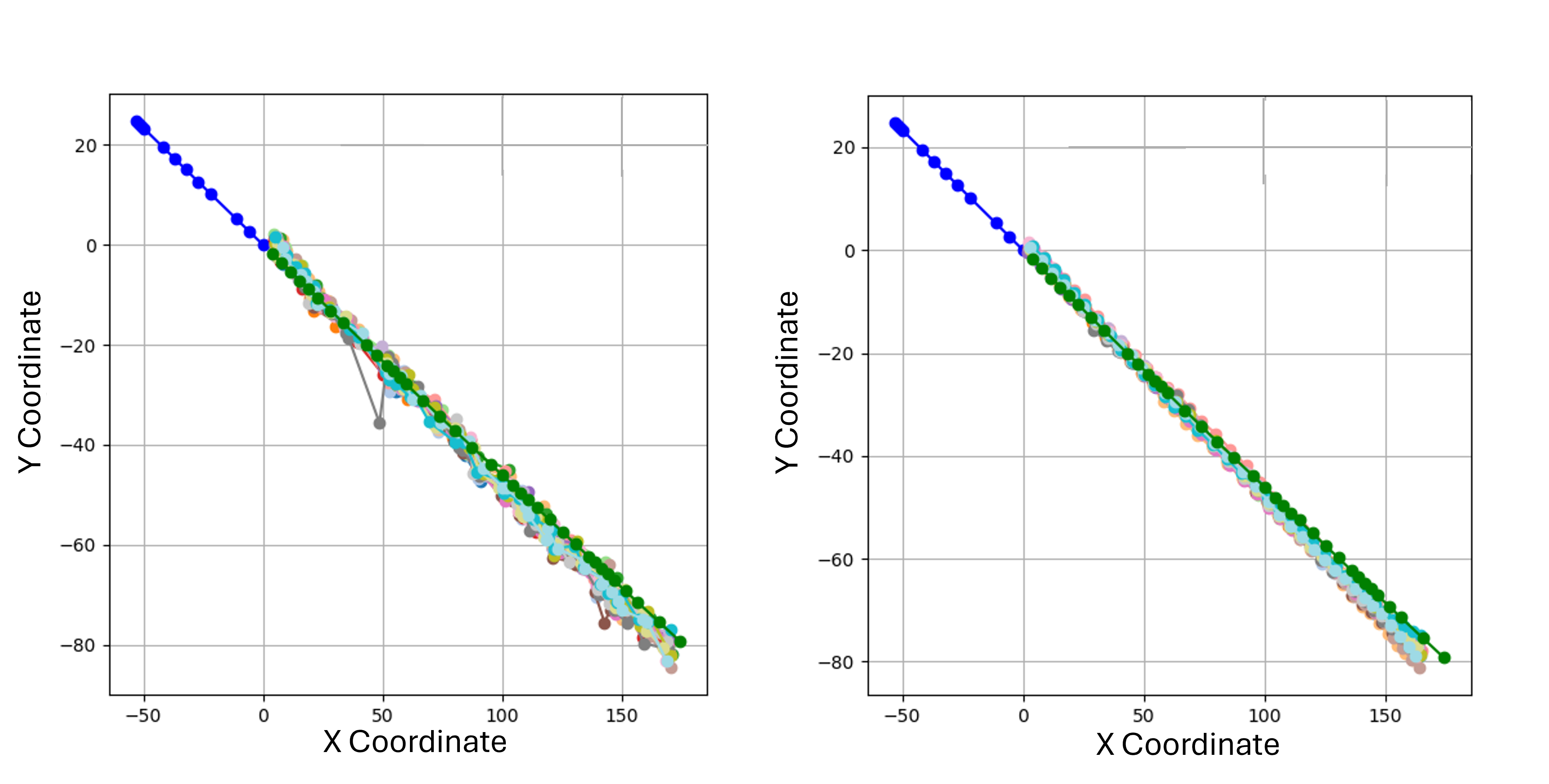}}\\[0.8em]
    \subfloat[]{\includegraphics[width=8.5cm]{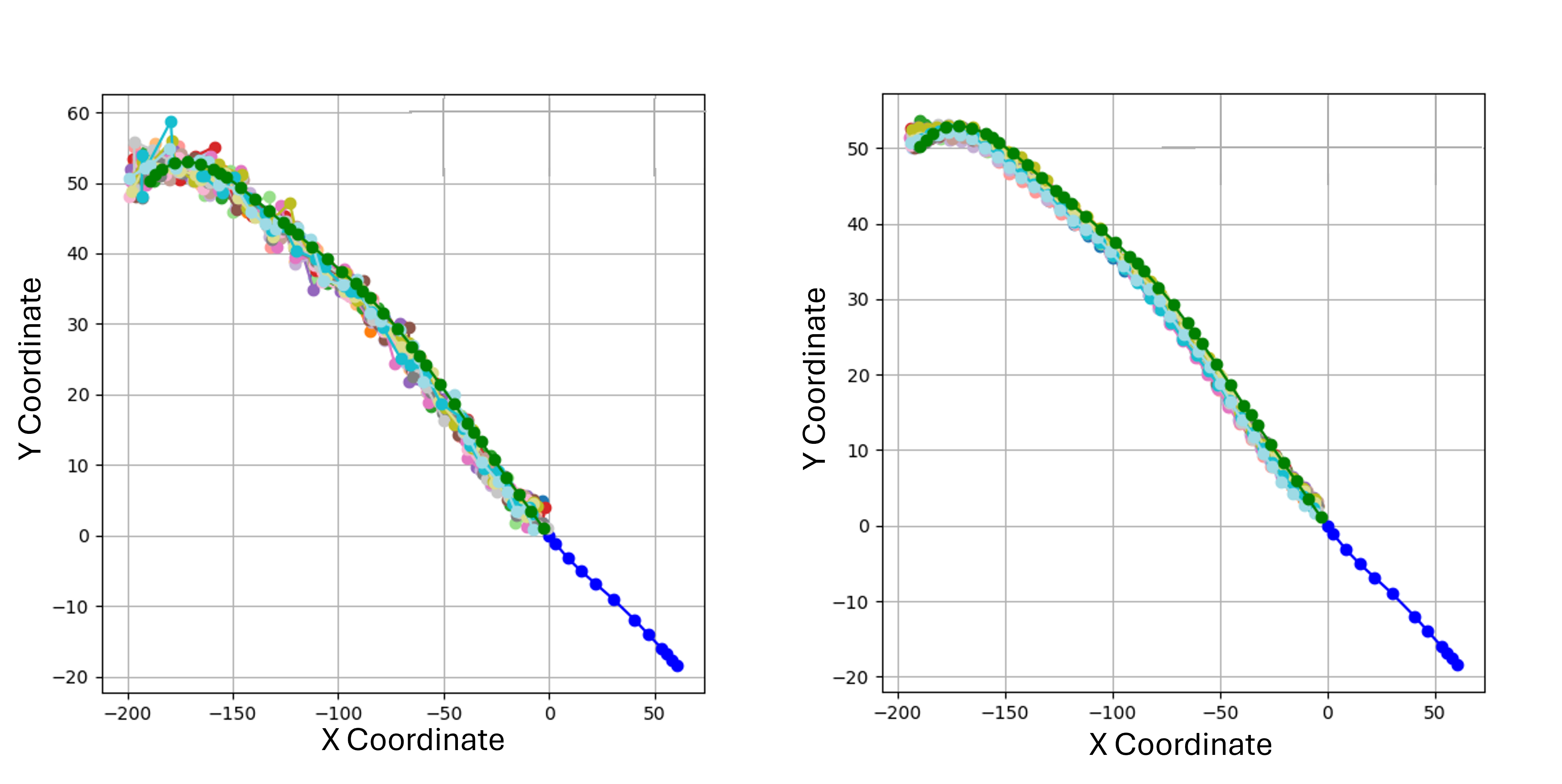}}\\[0.8em]
    \subfloat[]{\includegraphics[width=8.5cm]{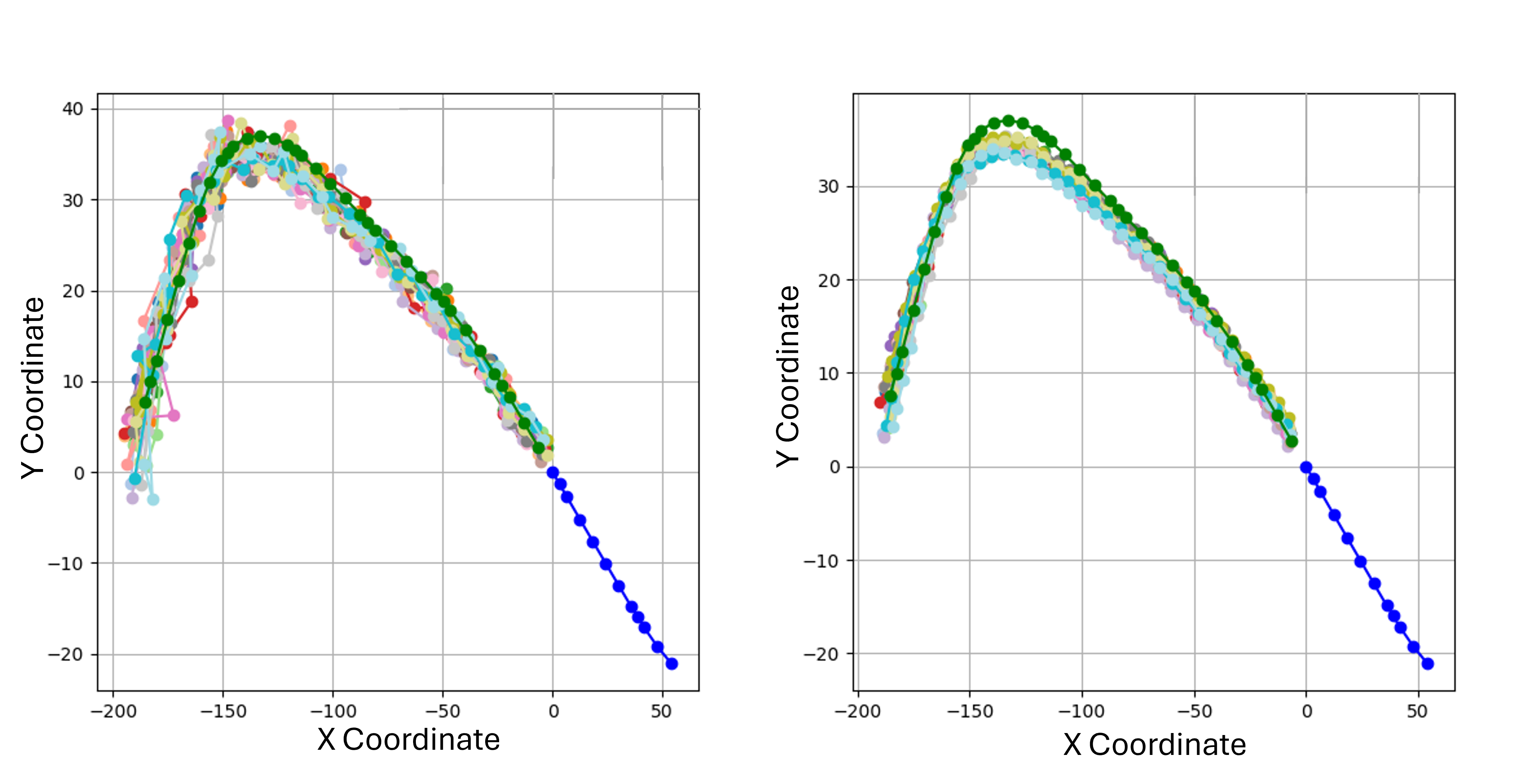}}

    \captionsetup{justification=justified}
    \caption{Three representative examples. In each row, the left plot shows the output prior to smoothing, and the right plot shows the corresponding output after applying the learned Savitzky-Golay filter.}
    \label{fig:ICM_2}
\end{figure*}

This phase of the ablation study assesses the effect of integrating the learnable Savitzky-Golay filter by comparing the performance of an ablation variant without a smoothing function with our proposed framework that includes the learnable Savitzky-Golay layer. The proposed framework with a learnable Savitzky–Golay layer demonstrates a superior performance compared to the variant that is without the smoothing function, as indicated in {Table~\ref{table:ICM_analysis}}. {Fig.~\ref{fig:ICM_2}} illustrates the variant with our proposed framework. The anticipated trajectories are superimposed on the ground reality in the illustration. {Fig.~\ref{fig:ICM_2}} demonstrates reduced high-frequency oscillations and a more piecewise-smooth curvature for the original framework with a learnable Savitzky–Golay layer, while preserving endpoint alignment and turn geometry. The analysis verifies that smoothing diminishes residual high-frequency components resulting from the interaction of short- and long-term predictors, while maintaining large-scale trajectory structure and final-point consistency.

\begin{table}[h]
\caption{Analysis on the importance of the Learnable SG filter.}
\centering
\begin{tabular}{|c|c|c|c|}
\hline
\textbf{Method} & \textbf{W/OUT Learnable SG filter} & \textbf{With Learnable SG filter}  \\ \hline
\textbf{ADE}    & 2.41        & 2.23       \\ \hline
\textbf{FDE}    & 3.81       & 3.68         \\ \hline
\end{tabular}
\label{table:ICM_analysis}
\end{table}

\subsection{Analysis on the effect of Unified Environment module}\label{ablation_cluster}

This phase of the ablation study assesses the effect of integrating the unified environmental module by comparing the performance of our proposed framework with two alternative forms of the framework. The two variations are: (i) No-Env (trajectory prediction devoid of environmental inputs) and (ii) Env-StaticConcat (environmental information concatenated with trajectory features, lacking learnt cross-modal interactions). The data presented in {Table~\ref{table:ENV_ablation}} demonstrates that the incorporation of environmental features reduces ADE and FDE compared to No-Env, suggesting that the availability of oceanic and meteorological influences enhances data-driven motion predictions. The proposed framework configuration results in further reductions in ADE and FDE compared to Env-StaticConcat. Our proposed framework enhances accuracy beyond static concatenation by (i) calculating attention over environmental tensors to emphasize regions and timestamps with greater causal impact on motion and (ii) conditioning the fusion on local vessel state and neighborhood context, allowing fusion weights to fluctuate based on heading, speed, and local traffic. These mechanisms synchronize environmental influences with the changing trajectory state, resulting in reduced ADE and FDE error measures.

\begin{table}[h]
\caption{Analysis on the importance of the unified environment module.}
\centering
\begin{tabular}{|c|c|c|c|}
\hline
\textbf{Method} & \textbf{No-Env} & \textbf{Env-StaticConcat} & \textbf{Ours}  \\ \hline
\textbf{ADE}    & 2.65        & 2.51 & 2.23      \\ \hline
\textbf{FDE}    & 4.07       & 3.90  & 3.68       \\ \hline
\end{tabular}
\label{table:ENV_ablation}
\end{table}

\subsection{Analysis on the effect of hierarchical stage-wise prediction}\label{ablation_cluster}

This section of the ablation study evaluates the impact of the two-stage design and its fusion mechanism by comparing four variants to the original framework using hierarchical gating fusion, all with consistent data, preprocessing, optimization, and training protocols. The four variants are: (i) a standard long-term-only model designed to predict the full horizon (Long-Only); (ii) a deeper, parameter-heavy long-term predictor scaled to match the capacity of the proposed framework (Capacity-Matched Long-Only); (iii) a two-stage model utilizing the element-wise sum of short- and long-term outputs (Sum-Fusion); and (iv) a two-stage model that concatenates the outputs, followed by a shallow MLP (MLP-Fusion).

When comparing single-stage baselines to two-stage fusion, ADE and FDE decrease dramatically. To ensure that performance increases are not primarily due to greater model capacity, we compare the two-stage models to the Deep Long-Only variation. While the conventional Long-just model has around 1.18 million parameters, expanding its depth to 2.04 million parameters (Capacity-Matched Long-Only) results in just a smaller improvement in ADE and a performance degradation in FDE. This degradation shows that just increasing the depth of a global attention network causes it to overfit to long-range priors rather than successfully capturing fine-grained local dynamics. This demonstrates that the two-stage framework outperforms a single long-term predictor via architectural disentanglement rather than parameter scaling alone. The long-term branch preserves global trajectory context and trends, while the short-term branch detects current kinematic changes and local environmental variations.

Furthermore, the findings highlight the importance of the unique hierarchical fusion architecture over naive combination tactics. The Sum-Fusion, MLP-Fusion, and the proposed model have nearly equal parameter counts (2.11M), as the fusion layers add a small footprint relative to the dual backbone. Despite the shared capacity, the original design with hierarchical gating fusion has the lowest ADE and FDE among all configurations.

The hierarchical fusion approach achieves optimal outcomes by performing data-dependent arbitration among the branches at each stage, conditioning the mixing weights on the current state and context. When immediate dynamics are clear, the gate increases reliance on short-term information; when ambiguity or noise is high, it prioritizes the long-term branch to stabilize the forecast. This context-aware mixture of experts reduces error accumulation and synchronizes each branch's contribution with its temporal dependability.

\begin{table}[h]
\caption{Analysis on the importance of hierarchical stage-wise prediction and model capacity.}
\centering
\resizebox{\columnwidth}{!}{%
\begin{tabular}{|l|c|c|c|c|c|}
\hline
\textbf{Method} & \textbf{Long-Only} & \textbf{Capacity-Matched Long-Only} & \textbf{Sum-Fusion} & \textbf{MLP-Fusion} & \textbf{Ours} \\ \hline
\textbf{Params (M)} & 1.18 & 2.04 & 2.11 & 2.11 & 2.11 \\ \hline
\textbf{ADE (km)}   & 4.50 & 4.38 & 2.92 & 2.77 & 2.23 \\ \hline
\textbf{FDE (km)}   & 6.11 & 6.70 & 4.99 & 4.39 & 3.68 \\ \hline
\end{tabular}%
}
\label{table:ENV_ablation}
\end{table}

\subsection{Sensitivity Analysis on Grid Granularity}\label{Sensitivity Analysis}

We performed a sensitivity analysis to evaluate the effect of spatial discretization on predictive performance by altering the number of grids $G$ in the short-term prediction module. Three configurations were assessed: $G=60$, $G=120$ (suggested), and $G=240$. Table VI illustrates that the $G=120$ configuration achieves the ideal equilibrium between spatial resolution and data density.

At $G=60$, the spatial cells do not possess the requisite resolution to reliably record the intricate, localized motion patterns and dynamics vital for good long-term forecasting. At $G=240$, the decrease in observation density inside each grid cell yields declining returns and may cause overfitting in sparsely observed seas, resulting in heightened displacement errors. The empirical results endorse the use of 120 grids as the standard for recording spatially changing motion fields while maintaining shared representation learning.

\begin{table}[h]
\caption{Sensitivity Analysis of Grid Resolution on ADE and FDE.}
\label{table:grid_sensitivity}
\centering
\begin{tabular}{|c|c|c|}
\hline
\textbf{Number of Grids ($G$)} & \textbf{ADE (km)} & \textbf{FDE (km)} \\ \hline
60 (Coarse) & 2.58 & 3.98 \\ \hline
\textbf{120 (Proposed)} & \textbf{2.23} & \textbf{3.68} \\ \hline
240 (Fine) & 2.41 & 3.85 \\ \hline
\end{tabular}
\end{table}

\subsection{Counterfactual Analysis of the Impact of Environmental Factors}

We conducted a controlled counterfactual analysis to assess the physical interpretability of the Unified Environmental Integration Module and to verify its sensitivity to the temporal correlation of oceanographic vectors. We conducted a comparison between the originally proposed framework and a variant model, in which the environmental input sequence was randomly shuffled along the temporal dimension. This experiment aims to isolate the module's response to dynamic forcing, thereby confirming that the cross-modal attention and FiLM layers effectively map the chronological evolution of physical sea states to vessel kinematics.

Empirical evaluation indicates a significant decline in prediction accuracy for the variant model. The application of environmental forces temporally misaligned with the vessel's actual conditions results in a misrepresentation of the temporal motion pathway, causing significant structural deviation from the ground truth. The induced kinematic error indicates that the framework fundamentally depends on the accurate temporal sequencing and magnitude of oceanographic forces to dictate the spatial evolution of the trajectory.

\begin{figure*}[t]
    \centering

    % Sub-figures side by side
    \subfloat[]{\includegraphics[width=0.3\textwidth]{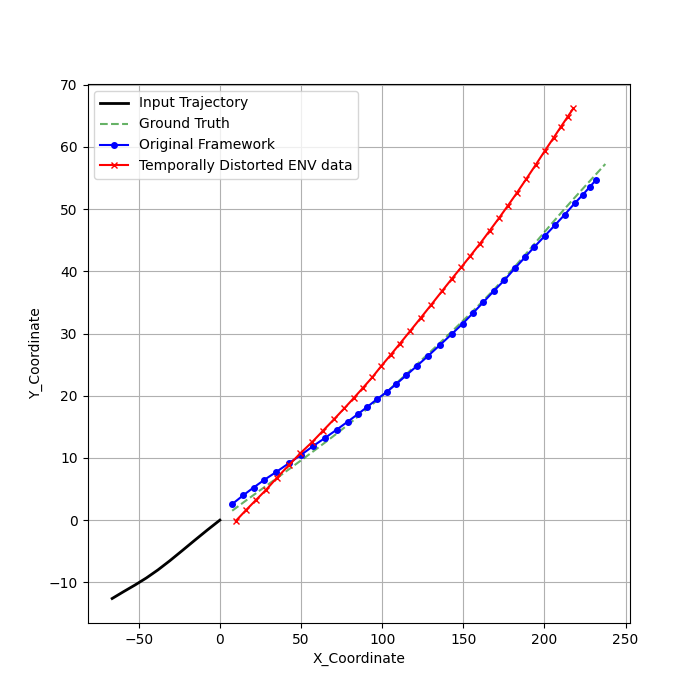}}
    \hspace{0.02\textwidth}
    \subfloat[]{\includegraphics[width=0.3\textwidth]{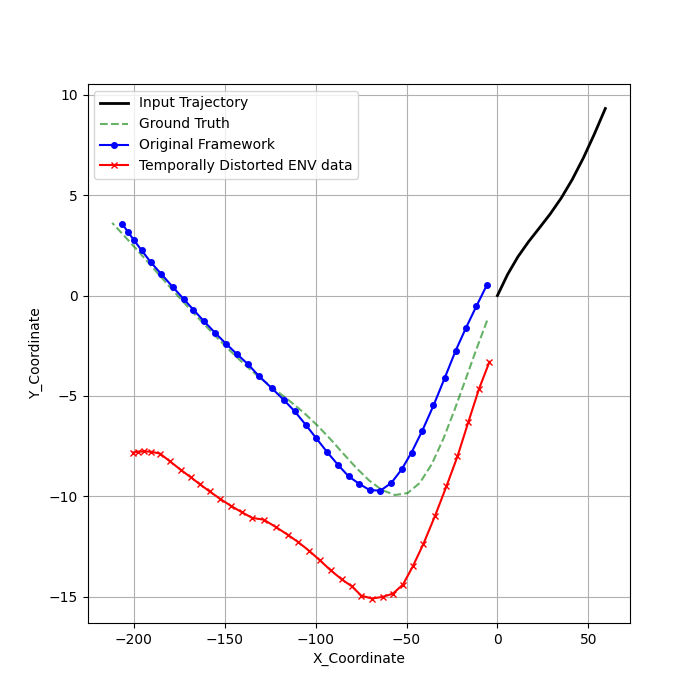}}
    \hspace{0.02\textwidth}
    \subfloat[]{\includegraphics[width=0.3\textwidth]{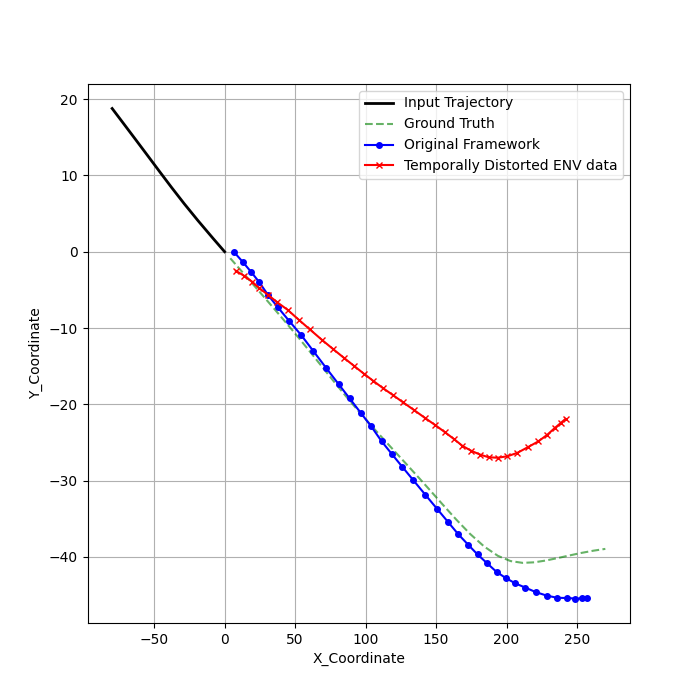}}

    \captionsetup{justification=justified}
    \caption{Three representative examples. In each plot, the blue line shows the output from the original framework, and the red line shows the output after temporally shuffling the environmental data.}
    \label{fig:causal_examples}
\end{figure*}

Figure~\ref{fig:causal_examples} illustrates this sensitivity by comparing the predictive outputs of both models in three distinct navigational scenarios. The variant model consistently produces deviations that are physically plausible yet structurally inaccurate given the vessel's actual timeline. The observed deviations significantly contrast with the precise spatial mapping of the originally proposed framework, thereby confirming the module's ability to explicitly interpret temporally coherent, vector-based physical forcing.

\subsection{Cross-Regional Spatial Generalization Analysis}\label{Cross-Regional Spatial Generalization Analysis}

This phase of the ablation study examines the framework's portability and addresses concerns about region-specific inductive biases by assessing cross-regional spatial generalization using a stringent stress test. We dynamically partitioned the domain by ranking the 120 geographical grids according to historical AIS ping density. To assess the model's ability to extrapolate from simple to complicated kinematic situations, the 100 lowest-density grids, consisting of sparse, open-ocean transit routes, were designated solely for the training set. The last 20 highest-density grids, including intricate shoreline approaches, port areas, and zones of intense interaction, constituted the designated test set. Throughout the training process, the shared Spatio-Temporal Graph Transformer backbone and hierarchical fusion module were exclusively optimized for the low-density source region. To assess portability, the backbone weights were fixed, and solely the 20 region-specific linear prediction heads were initiated and refined utilizing a limited selection of target-region data. We evaluated the proposed framework alongside the MSTFormer and Informer baselines using the same density-based spatial partitioning.

The comparison results in Table \ref{tab:cross_regional} indicate that the suggested framework attains enhanced spatial generalization, resulting in an Average Displacement Error (ADE) of 3.25 km and a Final Displacement Error (FDE) of 5.10 km. Although all models exhibit a predictable rise in error when shifting from sparse to very intricate interaction zones without complete retraining, our model preserves markedly superior structural integrity. This resilience is due to the architectural separation of global aim from local dynamics. The shared backbone acquires essential, transferable kinematics and foundational environmental reactions in open waters. The lightweight, grid-aware linear heads serve as an effective adaptation layer, swiftly adjusting the generalized latent features to the dense, highly nonlinear interaction patterns of the test grids. Conversely, the Informer and MSTFormer baselines demonstrate a significant decline in accuracy. Devoid of intricate interaction graphs during training, these global attention networks lack the localized adaptation mechanisms necessary to deduce complex evasive maneuvers and abrupt speed fluctuations in unfamiliar, high-traffic maritime areas, leading to increased compounding errors over extended durations.

\begin{table}[htbp]
\centering
\caption{ANALYSIS ON CROSS-REGIONAL SPATIAL GENERALIZATION .}
\label{tab:cross_regional}
\begin{tabular}{|l|c|c|}
\hline
\textbf{Method} & \textbf{ADE (km)} & \textbf{FDE (km)} \\
\hline
MSTFormer & 4.52 & 6.80 \\ \hline
Informer & 4.15 & 6.35 \\  \hline
\textbf{Ours} & \textbf{3.25} & \textbf{5.10} \\
\hline
\end{tabular}
\end{table}

\section{Conclusion}

The study focused on long-horizon vessel trajectory forecasting by developing a hierarchical two-stage architecture that integrates a coarse long-term predictor with grid-aware short-term stacks through a hierarchical fusion mechanism. The framework incorporates a unified environmental module that combines cross-attention with FiLM conditioning to integrate motion, spatial, and oceanographic signals. It also utilizes a learnable Savitzky–Golay smoothing layer to improve temporal coherence while maintaining geometric fidelity. Utilizing a standardized protocol (3 h input, 10 h prediction) on AIS data from the north west region of Australia, the proposed method demonstrated consistent enhancements in ADE/FDE compared to robust baselines. The inclusion of the environmental pathway yielded further improvements, highlighting the significance of oceanographic context in multi-hour planning.

The hierarchical design promotes global consistency and local fidelity but introduces capacity–granularity trade-offs in selecting and sizing grid-aware short-term stacks. This necessitates precise spatiotemporal alignment of environmental products. The proposed approach provides a highly accurate deterministic baseline. However, the lack of probabilistic safety margins for critical marine operations constitutes a significant limitation. Ablation studies show that learned fusion and gating reduce compounding errors compared to single-stage or naive blending strategies. However, their effectiveness can vary in rare maneuver regimes and sparsely observed waters. Additionally, the learnable smoothing layer relies on globally shared mixture weights, which may not be optimal across all sea states or vessel classes.

To overcome the above mentioned limitations promising future research directions encompass: (i) region-adaptive scaling of grid granularity to optimize accuracy and efficiency across diverse maritime environments; (ii) integration of enhanced exogenous drivers and more frequent ocean analyses within the unified environmental framework and (iii) uncertainty quantification closely linked to the fusion module to facilitate calibrated, risk-aware decision support for collision avoidance and route optimization.

\section*{Acknowledgement}
  
This work was supported by an Australian Research Council (ARC) Discovery research grant DP250103634.

\bibliographystyle{IEEEtran}
\bibliography{sample}
\begin{IEEEbiography}[{\includegraphics[width=1in,height=1.25in,clip,keepaspectratio]{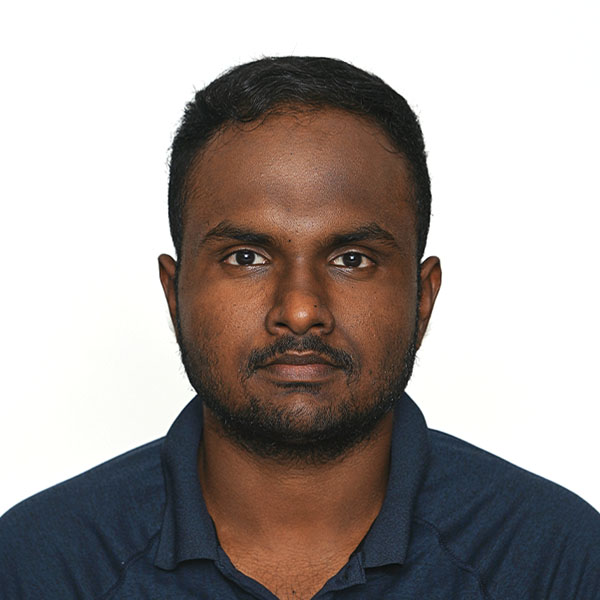}}]{Ganeshaaraj}
received the B.S. degree in Biomedical Engineering from the University of Moratuwa, Srilanka in 2022. He is currently pursuing a PhD degree
in Artificial Intelligence at Queensland University of Technology, Australia. His research
interests include computer vision, machine learning, artificial intelligence, signal processing, and medical imaging.
\end{IEEEbiography}
\begin{IEEEbiography}[{\includegraphics[width=1in,height=1.25in,clip,keepaspectratio]{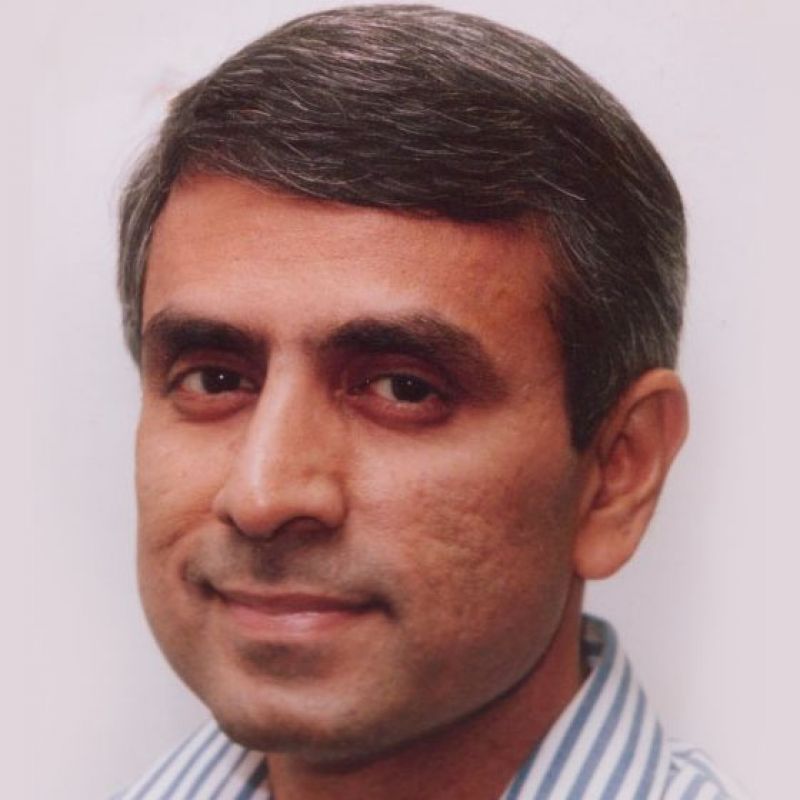}}]{Sridha Sridharan}
has obtained an MSc (Communication Engineering) degree from the University of Manchester, UK, and a PhD degree from the University of New South Wales, Australia. He is currently with the Queensland University of Technology (QUT) where he is a Professor in the School of Electrical Engineering and  Robotics. He has published over 600 papers consisting of publications in journals and in refereed international conferences in the areas of Image and Speech technologies during the period 1990-2023.  During this period he has also graduated 85  PhD students in the areas of Image and Speech technologies. Prof Sridharan has also received a number of research grants from various funding bodies including the Commonwealth competitive funding schemes such as the Australian Research Council (ARC) and the National Security Science and Technology (NSST) unit. Several of his research outcomes have been commercialised.
\end{IEEEbiography}
\begin{IEEEbiography}[{\includegraphics[width=1in,height=1.25in,clip,keepaspectratio]{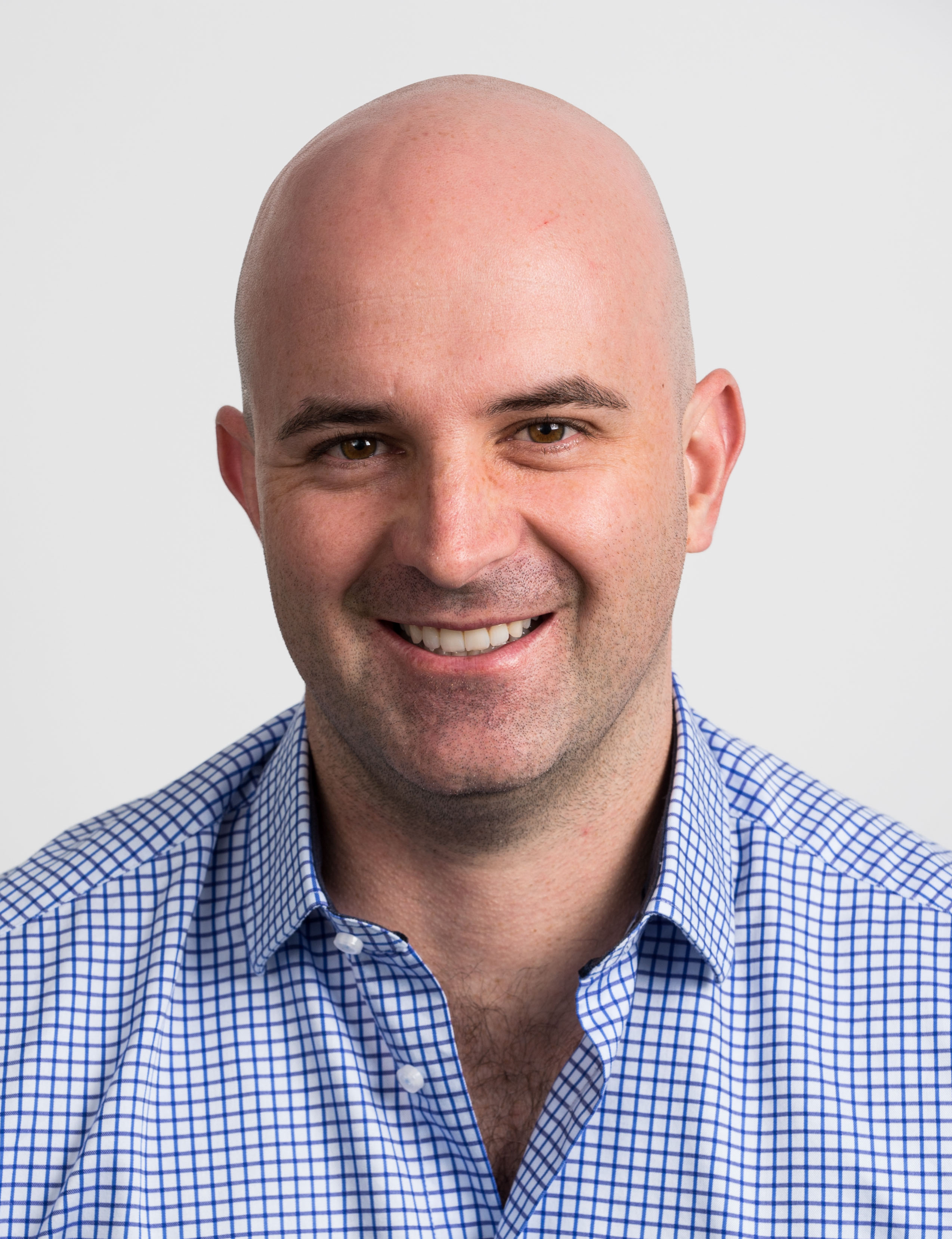}}]{Clinton Fookes}
is a Professor in Vision \& Signal Processing and is Co-Director of the SAIVT Lab at the Queensland University of Technology. He holds a BEng (Aero/Av), an MBA, and a PhD in computer vision. He has published over 350 internationally peer-reviewed articles and has attracted over \$35M of cash funding for fundamental and applied research from external competitive sources. This includes 11 Australian Category 1 grants funded from the Australian Research Council and the Department of Prime Minister and Cabinet. He serves on the editorial boards for the IEEE Transactions on Image Processing and Pattern Recognition, and has previously served on the editorial board for the IEEE Transactions on Information Forensics \& Security. He serves on the Asset Institute Board of Directors, the IEEE Biometrics Council, the IEEE Task Force on Deep Vision in Space, and is a representative on the Defence AI Key Stakeholder Group for the Defence Artificial Intelligence Research Network. He is a Senior Member of the IEEE, an Australian Institute of Policy and Science Young Tall Poppy, an Australian Museum Eureka Prize winner, Engineers Australia Engineering Excellence Awardee, Australian Defence Scientist of the Year, and a Senior Fulbright Scholar.
\end{IEEEbiography}
\begin{IEEEbiography}[{\includegraphics[width=1in,height=1.25in,clip,keepaspectratio]{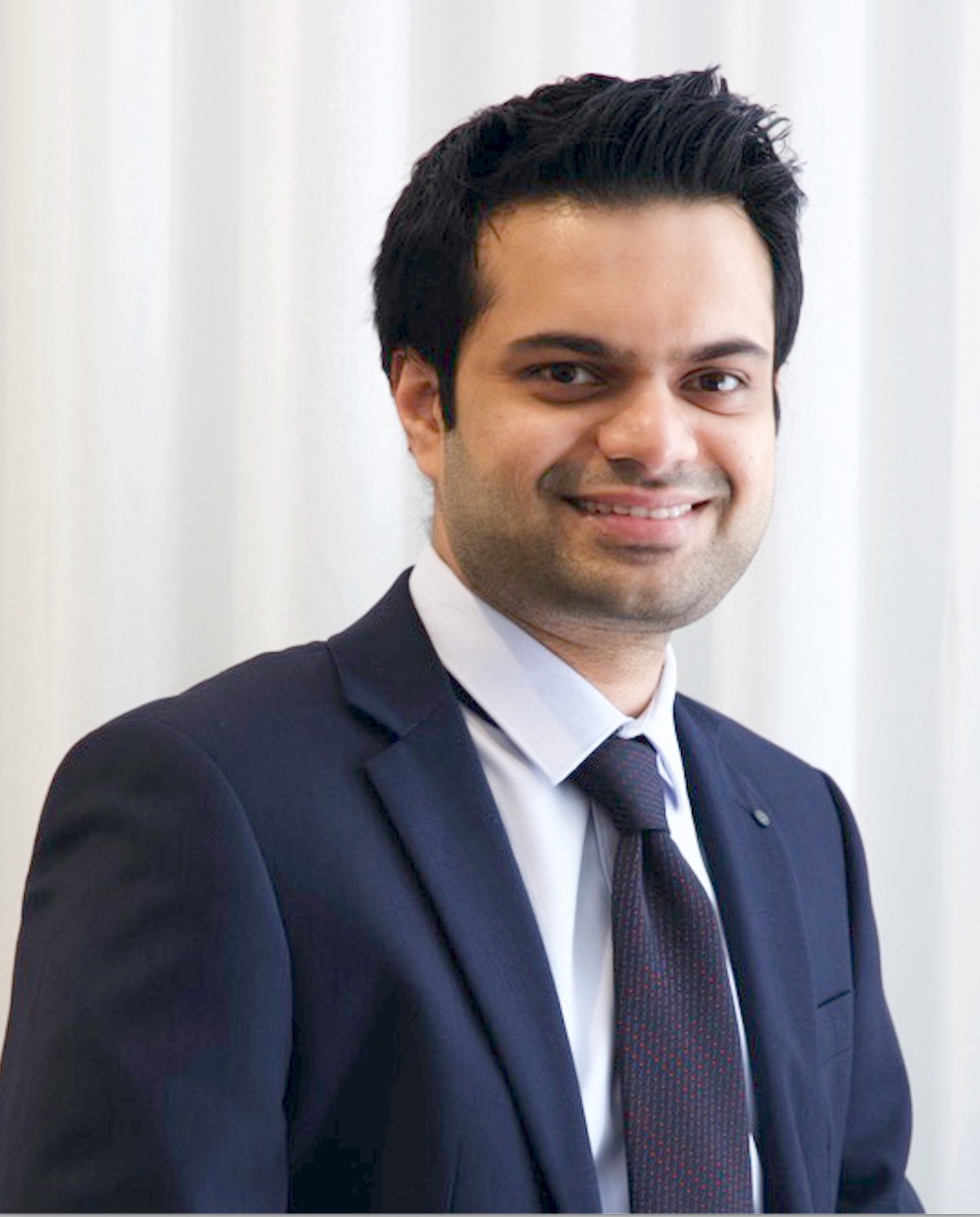}}]{Tharindu Fernando }
received his BSc (special degree in computer science) from the University of Peradeniya, Sri Lanka, and his PhD from Queensland University of Technology (QUT), Australia. He is currently a Postdoctoral Research Fellow in the Signal Processing, Artificial Intelligence, and Vision Technologies (SAIVT) research program at the School of Electrical Engineering and Robotics at Queensland University of Technology (QUT). He is a recipient of the 2019 QUT University Award for Outstanding Doctoral Thesis, the QUT Early Career Researcher Award in 2022, and the 2024 National Intelligence Post-Doctoral Grant. His research interests include Artificial Intelligence, Computer Vision, Deep Learning, Bio Signal Processing, and Video Analytics.
\end{IEEEbiography}

\end{document}